\documentclass{article}

  \PassOptionsToPackage{numbers, compress}{natbib}

  \usepackage[eandd,final]{neurips_2026}

  \usepackage[utf8]{inputenc} 
  \usepackage[T1]{fontenc}    
  \usepackage{hyperref}       
  \hypersetup{hidelinks}
  \usepackage[nottoc,notlot,notlof]{tocbibind} 

  \usepackage{url}            
  \usepackage{booktabs}       
  \usepackage{graphicx}       
  \usepackage{float}          
  \usepackage{wrapfig}        
  \usepackage{longtable}      
  \usepackage{tabularx}       
  \usepackage{multirow}       
  \usepackage{amsmath}        
  \usepackage{amsfonts}       
  \usepackage{amssymb}        
  \usepackage{nicefrac}       
  \usepackage{microtype}      
  \AtBeginDocument{{\itshape\global\rpcode\font`F=28\relax}}
  \usepackage[table]{xcolor}  
  \usepackage[most]{tcolorbox} 
  \usepackage{needspace}
  \usepackage{xurl} 
  \newcommand{\crmetric}[1]{Acc@{$\pm$}#1\%}

  \newcommand{\gmark}{\textcolor{green!55!black}{\ensuremath{\checkmark}}}
  \newcommand{\xmark}{\textcolor{black}{\ensuremath{\times}}}
  \newcommand{\appendixcontentssection}[3]{%
    \noindent\textbf{#1\quad #2}\dotfill\textbf{\pageref{#3}}\par\vspace{1.0em}}
  \newcommand{\appendixcontentssubsection}[3]{%
    \noindent\hspace{2.8em}#1\quad #2\dotfill\pageref{#3}\par\vspace{0.45em}}
  \newtcolorbox{kgpriorbox}{%
    colback=black!2,
    colframe=black!35,
    arc=3pt,
    boxrule=0.45pt,
    left=6pt,
    right=6pt,
    top=5pt,
    bottom=5pt,
    after={\par}}

  \title{Nereus\,\raisebox{-0.12em}{\includegraphics[height=0.95em]{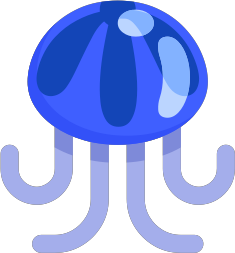}}: A Large-Scale Underwater Dataset for Fine-Grained Attribute Understanding and Grounded Counting Perception}

  \author{%
    \textbf{Jingtao Deng\textsuperscript{1,}}\thanks{{Both authors contributed equally to this research.}} \quad
    \textbf{Zihao Jin\textsuperscript{1,}}\footnotemark[1] \quad
    \textbf{Yuehe Chen\textsuperscript{1}} \\
    \textbf{Feixiang He\textsuperscript{3}} \quad
    \textbf{Shuiwang Li\textsuperscript{4}} \quad
    \textbf{Dan Zeng\textsuperscript{1,2,}}\thanks{{Corresponding author: {\texttt{zengd8@mail.sysu.edu.cn}}.}} \\[0.4em]
    {\normalfont\small\textsuperscript{1}School of Artificial Intelligence, Sun Yat-sen University{, Zhuhai, China}} \\
    {\normalfont\small{\textsuperscript{2}The Technology Innovation Center for Collaborative Applications of}} \\
    {\normalfont\small{Natural Resources Data in GBA, MNR, Guangzhou, China}} \\
    {\normalfont\small{\textsuperscript{3}School of Electronic Information, Central South University, Hunan, China}} \\
    {\normalfont\small{\textsuperscript{4}Guangxi Key Laboratory of Embedded Technology and Intelligent System,}} \\
    {\normalfont\small{Guilin University of Technology, Guilin, China}} \\[0.3em]
    {\normalfont\small\texttt{dengjt25@mail2.sysu.edu.cn}\quad\texttt{jinzh35@mail2.sysu.edu.cn}} \\
    {\normalfont\small\texttt{chenyh566@mail2.sysu.edu.cn}\quad{\texttt{drfxhe@gmail.com}}} \\
    {\normalfont\small{\texttt{lishuiwang0721@163.com}}\quad\texttt{zengd8@mail.sysu.edu.cn}}
  }

  \usepackage{etoolbox}
  \makeatletter
  \patchcmd{\@maketitle}{\@bottomtitlebar}{\@bottomtitlebar\vspace{-10pt}}{}{\errmessage{Title spacing patch failed}}
  \makeatother

\begin{document}
  \raggedbottom

  \maketitle

  \vspace{-2.1em}
  \begin{figure}[H]
    \centering
    \includegraphics[width=\linewidth,trim=24.00bp 161.50bp 19.00bp 116.50bp,clip]{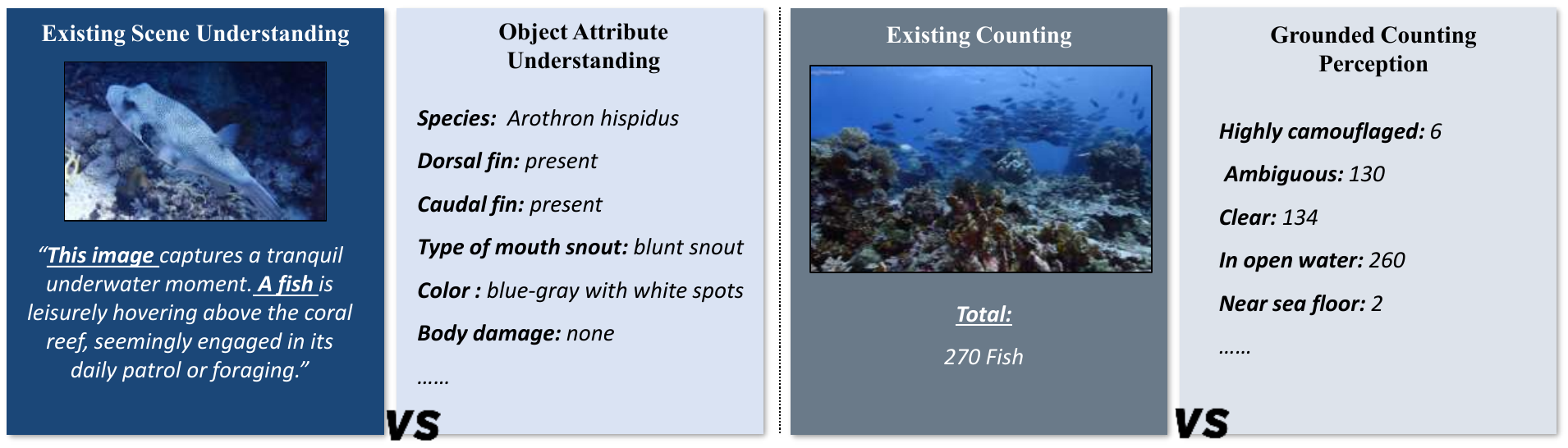}
    \vspace{-0.6em}
    \caption{Overview of Nereus. The dataset connects fine-grained organism attributes with grounded counting cues and degradation-aware underwater perception.}
    \label{fig:abstract-overview}
  \end{figure}
  \vspace{-0.5em}

  \begin{abstract}
  Underwater imagery records rich ecological information and supports applications such as science communication, fishery monitoring, ecological conservation, and oceanographic research.
  Beyond general scene understanding, these applications require models to recognize fine-grained biological traits and reason over spatially grounded evidence.
  However, existing underwater datasets often provide coarse category labels, generic captions, or total-count supervision, which limits their use for organism-level analysis and ecological reasoning.
  We introduce \textbf{Nereus}, a large-scale dataset for {Underwater Fine-grained Perception} that jointly supports Fine-Grained Object Attribute Understanding and Grounded Counting Perception.
  Nereus contains 88K images and 2.68M question-answer pairs, including annotations for 136K marine organism objects across 2,690 species and 1,515 genera, as well as 65K point annotations and 67K grounded counting QA pairs.
  Experiments with representative multimodal models show that Nereus improves downstream adaptation over existing underwater-domain data and provides a foundation for future research on fine-grained marine perception.
  {Our Nereus dataset and code are available at {\url{https://brl-sysu.github.io/Nereus/}}.}
  \end{abstract}

  \begin{table}[!t]
    \caption{Comparison between Nereus and existing underwater understanding datasets. Nereus contains 88K images, 2.68M {question-answer pairs}, 136K objects, and 65K point annotations, and mainly supports two tasks highly related to marine downstream applications: Fine-Grained Object Attribute Understanding and Grounded Counting Perception.}
    \label{tab:dataset_comparison}
    \centering
    \normalsize
    \resizebox{\linewidth}{!}{
    \begin{tabular}{@{}lllcccccccc@{}}
      \toprule
      \multirow{2}{*}[0.3ex]{Datasets} & \multirow{2}{*}[0.3ex]{Venue} & \multirow{2}{*}[0.3ex]{Img} & \multicolumn{3}{c}{Annotation Modality} & \multicolumn{5}{c}{Supported tasks} \\
      \cmidrule(lr){4-6} \cmidrule(lr){7-11}
      & & & \begin{tabular}[c]{@{}c@{}}Object-level Attr.\\Annotation\end{tabular} & QA Pair & Instance Desc. & Detection & VQA & Caption & \begin{tabular}[c]{@{}c@{}}Fine-Grained Attr.\\Understanding\end{tabular} & \begin{tabular}[c]{@{}c@{}}Counting\\Perc.\end{tabular} \\
      \midrule
      MarineInst20M~{{\cite{marineinst20m}}} & ECCV 2024 & 2.42M & \xmark & \xmark & \gmark & \gmark & \xmark & \gmark & \xmark & \xmark \\
      NautData~{{\cite{nautilus}}} & NeurIPS 2025 & 158K & \xmark & \gmark & \gmark & \gmark & \gmark & \gmark & \xmark & \xmark \\
      Fish-Vista~{{\cite{fishvista}}} & CVPR 2025 & 69.3K & \gmark & \xmark & \xmark & \xmark & \xmark & \xmark & \gmark & \xmark \\
      CoralVQA~{{\cite{coralvqa}}} & NeurIPS 2025 & 12.8K & \gmark & \gmark & \xmark & \xmark & \gmark & \xmark & \gmark & \xmark \\
      UWBench~{{\cite{uwbench}}} & arXiv 2025 & 15.0K & \xmark & \gmark & \gmark & \xmark & \gmark & \gmark & \xmark & \xmark \\
      IOCFish5K~{{\cite{iocfish5k}}} & CVPR 2023 & 5.6K & \xmark & \xmark & \xmark & \xmark & \xmark & \xmark & \xmark & \gmark \\
      \midrule
      \rowcolor{orange!15}
      Nereus (Ours) &  & 88K & \gmark & \gmark & \gmark & \gmark & \gmark & \gmark & \gmark & \gmark \\
      \bottomrule
    \end{tabular}}
  \end{table}

  \section{Introduction}

  Underwater ecosystems contain rich biodiversity, complex ecological interactions, and valuable natural resources. Large-scale underwater imagery provides an important visual record for science communication, fishery monitoring, ecological conservation, and oceanographic research. Recent progress in computer vision and multimodal large language models (LMMs) makes it increasingly possible to extract ecological knowledge from such imagery and reduce manual observation costs. Unlike in-air visual perception, underwater scenes suffer from severe degradations and domain gaps, such as color cast, scattering, turbidity, blur, low contrast, and occlusion, making fine-grained visual perception highly challenging. {While} general underwater scene understanding has received increasing attention, fine-grained underwater visual perception remains unexplored.
  \begin{figure}[!b]
    \centering
    \includegraphics[width=\linewidth]{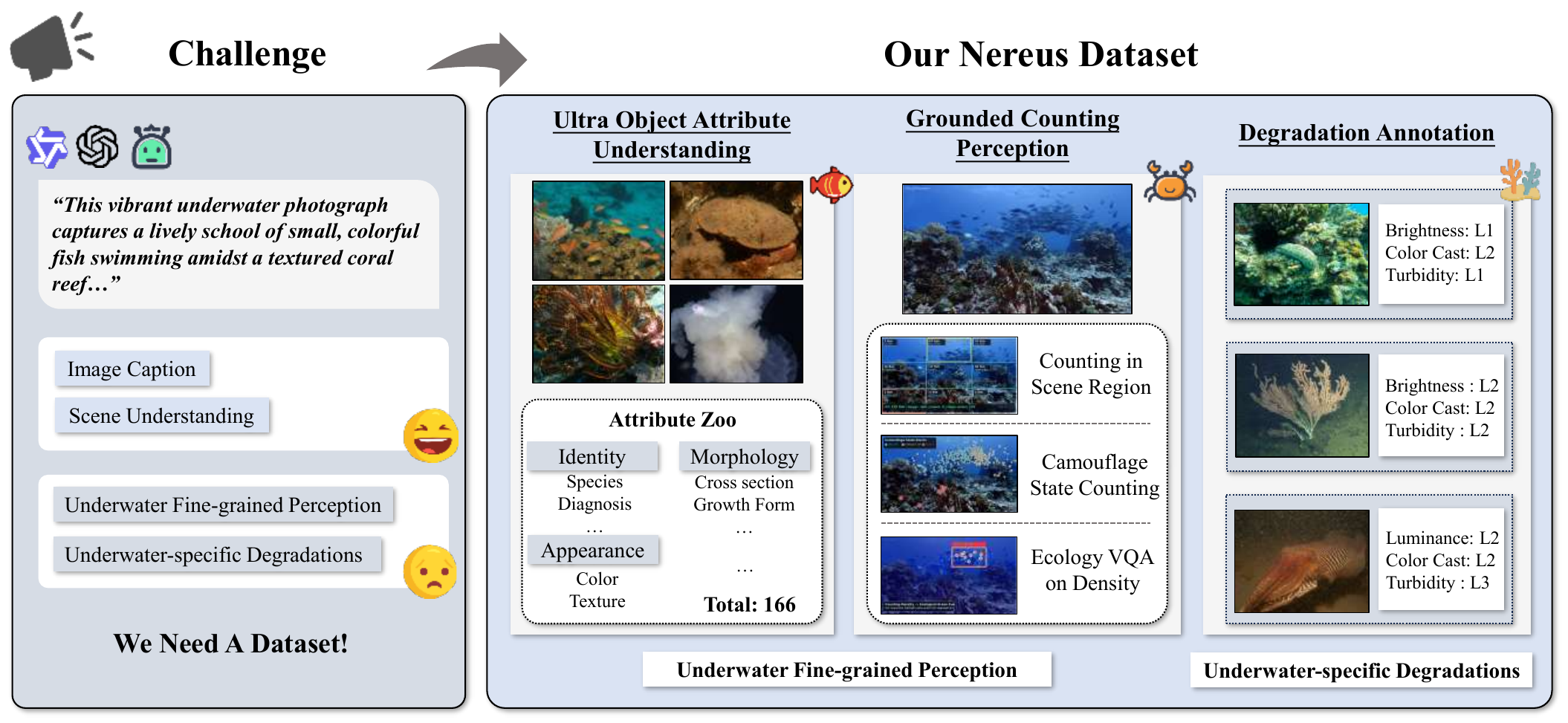}
    \caption{{Motivation of Nereus. The gap between underwater application needs and existing resources motivates a unified dataset for fine-grained object understanding, grounded counting, and degradation-aware analysis.}}
    \label{fig:intro-motivation}
  \end{figure}

  General underwater vision-language datasets mainly emphasize scene-level descriptions, captions, referring expressions, detection, or broad visual question answering. Taxon-specific datasets provide valuable biological supervision, but are usually restricted to a limited set of organism groups, such as fish or coral. Counting datasets provide point annotations or numerical supervision, yet they usually treat counting as an isolated task of estimating a total number rather than reasoning over semantic regions, visibility conditions, density patterns, or ecological context. As illustrated in Figure~\ref{fig:intro-motivation}, this leaves a gap between existing underwater resources and application-oriented underwater perception: current resources are relatively effective for image captioning and coarse scene understanding, but provide limited support for fine-grained visual perception including object-level biological attributes, grounded counting perception, and underwater-specific degradation analysis.

  Table~\ref{tab:dataset_comparison} further summarizes this gap from the perspectives of data scale, annotation modality, and supported tasks . Existing datasets differ in their strengths: MarineInst20M~\cite{marineinst20m} provides large-scale marine instance descriptions, NautData~\cite{nautilus} and UWBench~\cite{uwbench} support general underwater vision-language understanding, Fish-Vista~\cite{fishvista} and CoralVQA~\cite{coralvqa} contribute fine-grained supervision for specific taxonomic groups, and IOCFish5K~\cite{iocfish5k} provides point-level supervision for underwater fish counting. However, these resources do not jointly provide structured object-level attribute annotations, Question Answer supervision, descriptions, detection-related localization, captioning support, fine-grained attribute understanding, and counting perception within a unified dataset. In particular, existing counting resources rarely transform point annotations into region-aware, camouflage-aware, and density-grounded instruction data, which limits their use for fishery monitoring and ecological reasoning.

  {In this work, we focus on four underwater applications: Science Communication~{\cite{campbell2023identifiers}}, Fishery Monitoring~{\cite{langlois2020stereo}}, Ecological Conservation~{\cite{yamano2004bleaching}}, and Oceanographic Research~{\cite{costello2015traits}}.} These scenarios motivate \textbf{{Underwater Fine-grained Perception}}, which requires object-level biological attributes, spatially grounded counting evidence, and degradation-aware interpretation. We introduce \textbf{Nereus}, a large-scale underwater dataset organized around two task families: \textbf{Fine-Grained Object Attribute Understanding}, which provides cross-taxon object-level supervision over taxonomy, morphology, appearance, condition, pose, visibility, and organism-environment interaction; and \textbf{Grounded Counting Perception}, which extends counting from numerical estimation to region-aware, camouflage-aware, and density-grounded reasoning. All images are additionally annotated with luminance, color cast, and turbidity-related degradation levels.

  {Overall, Nereus contains 88K images, 2.68M question-answer pairs, 136K objects, and 65K point annotations. Our benchmark demonstrates the benefits of Nereus supervision while revealing remaining challenges in underwater perception.}
  Our contributions are summarized as follows:
  \begin{itemize}
      \item We introduce \textbf{{Underwater Fine-grained Perception}}, an application-oriented multimodal problem that goes beyond general underwater scene understanding. We define two core capabilities: \textbf{Fine-Grained Object Attribute Understanding} and \textbf{Grounded Counting Perception}, covering science communication, fishery monitoring, ecological conservation, and oceanographic research.

      \item {We build {\textbf{Nereus}}, a large-scale underwater dataset with finer-grained biological attributes and grounded counting supervision. Nereus contains 88K images, 2.68M question-answer pairs, 136K annotated marine organism objects, 136 attribute dimensions, 2,690 species, 1,515 genera, 65K point annotations, and 67K grounded counting QA pairs. Its annotations are further verified through LMM-assisted screening and human review.}

      \item We provide a comprehensive benchmark with representative open-source and {proprietary LMMs}, including {LLaVA, InternVL, Qwen, and Gemini}. Experiments show that Nereus improves downstream adaptation over prior underwater-domain data; for example, compared with NautData, it raises overall attribute accuracy {from 11.27\% to 51.44\% on LLaVA-OneVision, 15.13\% to 51.88\% on InternVL2.5, and 17.53\% to 46.17\% on Qwen2.5-VL.}
  \end{itemize}

\section{Related Work}
\subsection[General Underwater Scene Understanding]{\texorpdfstring{{General Underwater Scene Understanding}}{General Underwater Scene Understanding}}
Large multimodal models have advanced captioning, VQA, grounding, referring expressions, and multimodal reasoning~\cite{blip2,llavaov,qwen25vl,internvl25,gemini}, but underwater images introduce domain-specific degradations such as color cast, scattering, turbidity, blur, low contrast, and occlusion~\cite{seathru,uieb,uciqe,uiqm}. Underwater scene resources such as SUIM~\cite{suim} and recent marine multimodal datasets, including MarineInst20M~\cite{marineinst20m}, NautData~\cite{nautilus}, UWBench~\cite{uwbench}, CoralVQA~\cite{coralvqa}, Fish-Vista~\cite{fishvista}, and MarineGPT~\cite{marinegpt}, support segmentation, instance descriptions, instruction following, captioning, grounding, VQA, or taxon-specific understanding. These works are valuable, but they mainly focus on scene-level understanding, benchmark evaluation, instance description, or specific organism groups, leaving cross-taxon biological attributes and grounded counting perception under-supported.

\subsection[Fine-Grained Underwater Visual Perception]{\texorpdfstring{{Fine-Grained Underwater Visual Perception}}{Fine-Grained Underwater Visual Perception}}
{Underwater Fine-grained Perception requires models to answer questions about subtle biological and spatial evidence, such as visible organism traits, densest regions, camouflage states, and habitat-related density patterns. Recent studies show that progress on general multimodal benchmarks does not necessarily translate into fine-grained visual recognition~{\cite{ghosh2026finegrained}}. FG-BMK examines fine-grained semantic recognition and visual representations, identifying limitations in modality alignment and category-level reasoning~{\cite{yu2026fgbmk}}. Finedefics improves fine-grained recognition through object-attribute supervision and attribute-aware alignment~{\cite{he2025finedefics}}.} {BioCLIP2 learns biological visual representations through large-scale contrastive vision-language training, supporting species recognition and trait prediction~{\cite{gu2025bioclip2}}.}

General fine-grained resources such as CUB-200-2011~\cite{cub} and EagleVision~\cite{eaglevision} show the value of structured object attributes, while biological databases such as FishBase~\cite{fishbase}, WoRMS~\cite{worms}, and coral trait databases~\cite{coraltraitdb} provide rich trait knowledge. However, these resources are not directly aligned with underwater visual evidence. Existing underwater datasets provide only partial support through fish traits, coral VQA~\cite{coralvqa}, or general marine descriptions. Nereus complements them by connecting object-level biological attributes with region-aware counting, camouflage-aware counting, and density-hotspot reasoning.

\section{Nereus Overview}
  \label{sec:dataset}

  \begin{figure}[H]
    \centering
    \includegraphics[width=0.92\linewidth]{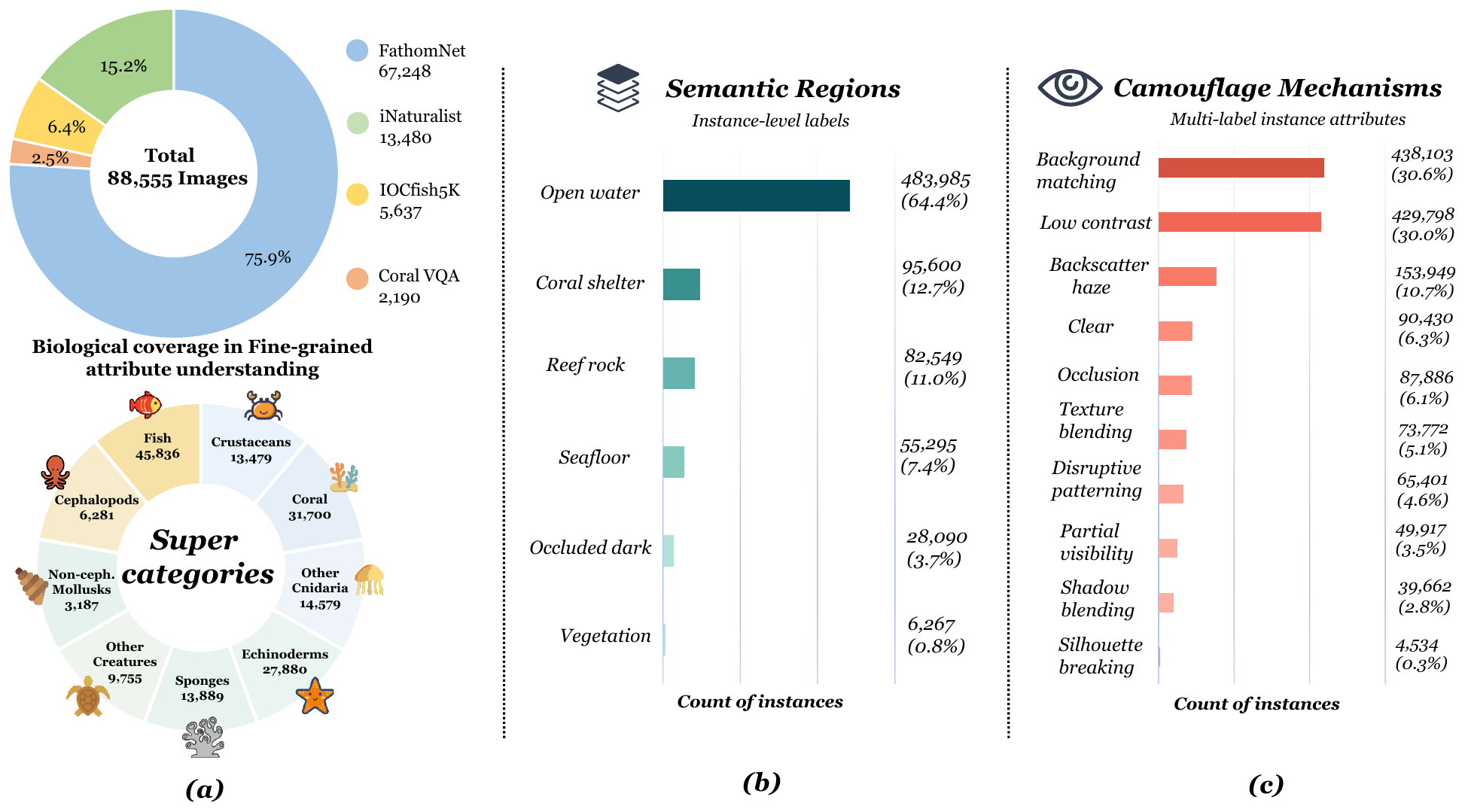}
    \caption{{Overview statistics of Nereus. Data-source composition and biological diversity for fine-grained object understanding, alongside semantic-region and camouflage-mechanism distributions for grounded counting.}}
    \label{fig:dataset-statistics}
  \end{figure}

  \subsection{Overview}
  \textbf{Data Curation.}
  Nereus is a large-scale underwater dataset for Fine-Grained Attribute Understanding and Grounded Counting Perception. It contains {88K images} in total and is split into training and test sets, with 86{,}376 images for training and 2{,}179 images with 16{,}744 QA pairs for model evaluation. We construct the test set by sampling according to the distribution of super categories and include species absent from the training set, allowing evaluation of cross-species generalization. We collect and integrate data from four sources: FathomNet~{{\cite{fathomnet}}}, CoralVQA~{{\cite{coralvqa}}}, iNaturalist~{{\cite{inaturalist}}}, and IOCFish5K~{{\cite{iocfish5k}}}. FathomNet serves as the primary source for object-level marine organism annotations, CoralVQA and iNaturalist complement coral, sponge, and fish data, and IOCFish5K provides point annotations for counting-oriented perception. The number of samples collected from each dataset is summarized in Figure~\ref{fig:dataset-statistics}. License information for the data sources and our released annotations is provided in Appendix~\ref{app:license-details}.

  \textbf{Statistics for Fine-Grained Object Attribute Understanding.}
  Following the biological distribution of FathomNet, we organize marine organisms into nine major groups: Fish, Coral, Cephalopods, Crustaceans, Echinoderms, Non-Cephalopod Mollusks, Other Cnidarians, Sponges, and Other Creatures~\cite{fathomnet}. Nereus contains object annotations across these groups, and the object-count composition is shown in Figure~\ref{fig:dataset-statistics}. To better support downstream applications, we define a 136-dimensional attribute set by consulting biological resources such as FishBase~{{\cite{fishbase}}}, Coral Trait Database~{{\cite{coraltraitdb}}}, and Fish-Vista~{{\cite{fishvista}}}, as well as by analyzing downstream application requirements. The attribute set covers five categories: identity, structural morphology, appearance and surface, condition and environment, and pose, visibility, and interaction, with an overview of the attribute taxonomy provided in Appendix Figure~\ref{fig:attribute-taxonomy}.

  \textbf{Statistics for Grounded Counting Perception.}
  Nereus supports three counting perception tasks: region-aware counting, camouflage-state counting, and density reasoning. To support these tasks, we define underwater scene types and instance-level camouflage states, so that counting supervision is connected with spatial context and visibility conditions. The proportions of different scenes and camouflage states are shown in Figure~\ref{fig:dataset-statistics}.

  \subsection{Annotation Quality and Accuracy}
  {The reliability of the VLM-generated annotations is supported by two complementary strategies. First, we use object-centered crops, biological knowledge-graph priors, and multi-model interaction for automatic annotation, as described in Section~{\ref{sec:attribute-construction}}. Second, after automatic annotation, human experts review and correct a sampled subset. Human participation is central to our annotation quality control. We first design a 10-point rating scale and use GPT-4o to score a small randomly sampled subset of Nereus annotations~{\cite{gpt4o}}. Three master's students in marine biology manually review annotations, correct errors and insufficient content, and calibrate the 10-point scale for LLM-assisted annotation quality assessment. Details of the expert review, inter-annotator agreement, and annotation error rate are provided in Appendix~{\ref{app:expert-review-iaa}}.} The calibrated 10-point scale, its four-level human-verification mapping, and the semantic-similarity scale are provided in Appendix Tables~\ref{tab:10-point-scale}--\ref{tab:100-point-sim-scale}; these criteria guide test-set scoring and expert arbitration to ensure annotation accuracy.

  \textbf{{Training-Set Evaluation.}}
  We randomly sample 200 images from the training set and evaluate them with a four-level quality scale. The training annotations obtain an average score of 9.26, supporting the reliability of Nereus annotations.

  \textbf{{Test-Set Evaluation.}}
  Using the 10-point scale, we score all annotations in the Nereus test set. The annotations achieve an average score of 9.14, and 94.8\% receive scores of 8 or above. {We further submit annotations with scores below 8 to domain experts; in total, 63 objects are manually arbitrated. Among the 934 reviewed attributes, 43 are corrected, corresponding to a correction rate of 4.60\%.}

  \section{Construction}

  \begin{figure}[!t]
    \centering
    \includegraphics[width=\linewidth,trim=29.00bp 101.00bp 23.50bp 81.50bp,clip]{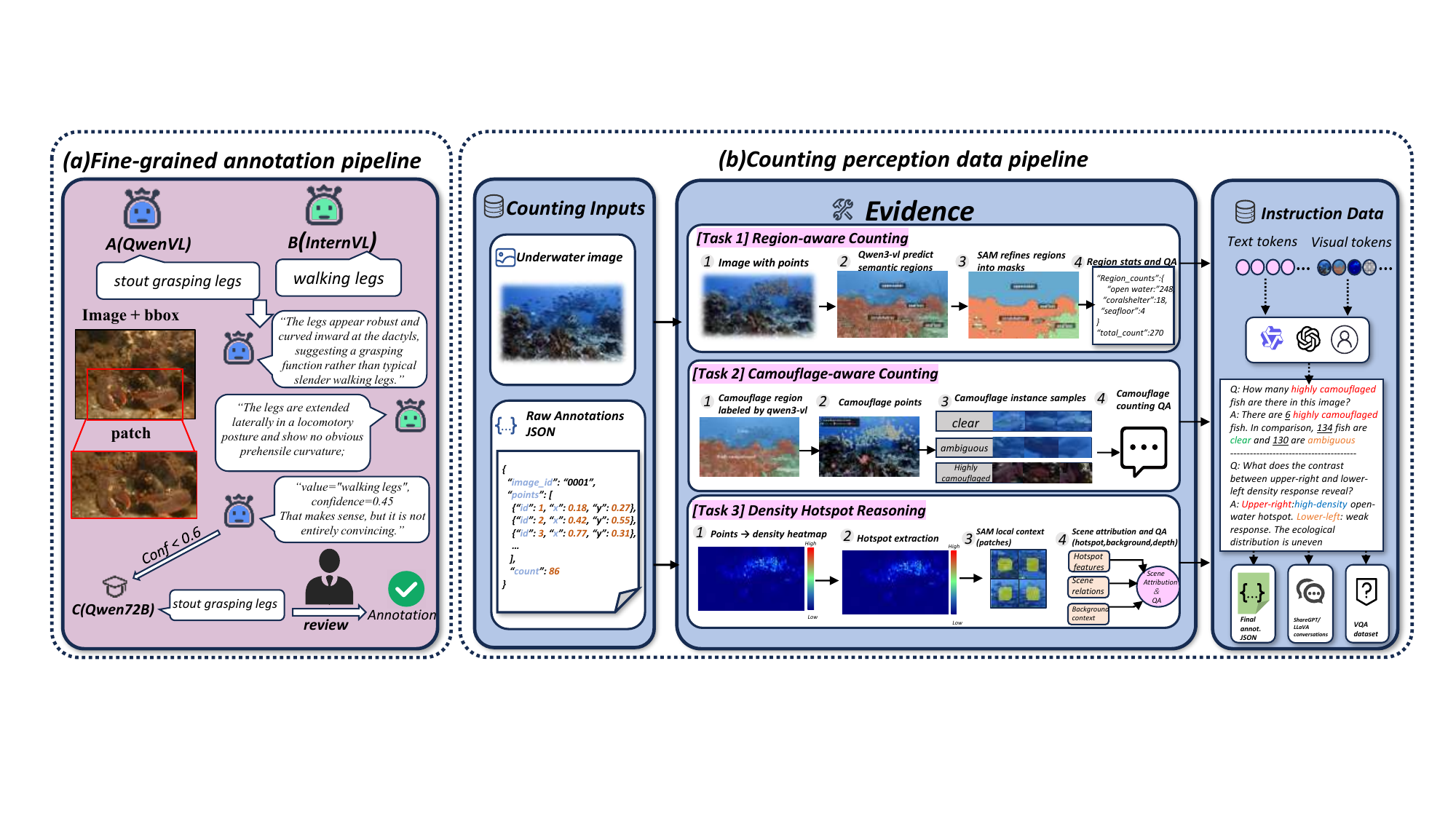}
    \caption[Construction pipeline of Nereus.]{{{\textbf{Construction pipeline of Nereus.}} (a) {\textbf{Fine-grained annotation pipeline:}} Multi-model discussion and arbitration generate structured biological attributes using object crops, spatial context, and biological priors. (b) {\textbf{Counting perception data pipeline:}} Point annotations are transformed into region-aware, camouflage-aware, and density-grounded QA pairs.}}
    \label{fig:construction-pipeline}
  \end{figure}

  \subsection{Fine-Grained Object Attribute Understanding}
  \label{sec:attribute-construction}
  In this section, we describe how Nereus constructs Fine-Grained Object Attribute Understanding annotations from heterogeneous underwater images. Given the predefined attribute set described in Section~\ref{sec:dataset}, the pipeline focuses on preparing object-centered visual inputs, injecting biological knowledge priors, and producing reliable annotations through multi-model collaboration and arbitration.

  \textbf{{Object-Centered Preprocessing.}}
  Following EagleVision, we mainly process each annotated organism into an object-centered input so that the annotation model can focus on the target itself~\cite{eaglevision}. For data without bounding boxes or center-point annotations (iNaturalist~{{\cite{inaturalist}}}), we use GroundingDINO~{{\cite{groundingdino}}} to generate bounding boxes and randomly sample 20\% of them for manual review and correction. For data with center-point annotations (CoralVQA), manual review shows that almost all sampled images contain a single dominant subject, which is sufficient for the model to focus on the target organism. After obtaining the target localization, we crop an object patch around the bounding box. We expand the crop with a 15\% margin, which preserves nearby visual cues that are useful for interaction-related attributes. More processing details are provided in Appendix~\ref{app:data-processing-details}.
  Also, underwater images often suffer from color cast, turbidity, low contrast, and blur, making local object patches insufficient for reliable fine-grained attribute annotation. To address this challenge, we provide the original image together with bounding box coordinates or point annotations as global context, allowing the annotation model to use broader scene semantics when local visual evidence is degraded. We also introduce degradation annotations for all images. Specifically, based on the underwater image quality index (UIQI), we divide underwater degradation into three types and three severity levels, and annotate each image accordingly~\cite{uiqi}. Detailed metric definitions, processing procedures, and level partitioning criteria are provided in Appendix~\ref{app:degradation-details}.

  \textbf{{Knowledge Graph Construction.}}
  To improve annotation accuracy, we construct a knowledge-graph prior from FishBase~{{\cite{fishbase}}}, WoRMS~{{\cite{worms}}}, Wikipedia~{{\cite{wikipedia}}}, Coral Trait Database~{{\cite{coraltraitdb}}}, and other biological databases. The knowledge graph contains trait information and textual descriptions for 2{,}803 species or genera, covering taxonomy, morphology, appearance, habitat, and ecological traits. {During annotation, we retrieve the corresponding entry according to each object's concept name and inject the prior knowledge into the annotation prompt, which explicitly instructs the model to prioritize visible image evidence over the retrieved prior knowledge.} An example prior prompt and its final annotation fields are provided in Appendix~\ref{app:kg-prior-example}.

  \textbf{{Automated Annotation.}}
  We conduct pilot annotation on the test set and select the best-performing VLM combination~\cite{vlmcombination} from candidate models: Qwen3-VL-8B~{{\cite{qwen3}}} and InternVL3-8B~{{\cite{internvl3}}}. We design a ``dual-model discussion with single-model arbitration'' annotation pipeline. Specifically, model A (Qwen3-VL-8B) and model B (InternVL3-8B) independently annotate each input based on the patch image, original image with bounding box coordinates, attribute definitions, and knowledge-graph prior. The two models then enter a three-turn discussion. In Turn 1, model A reviews model B's answers and provides agreement decisions and reasons for each attribute. In Turn 2, model B updates its answers according to model A's feedback. In Turn 3, model A integrates the full discussion and outputs the final consensus value and agreement confidence for each attribute. During arbitration, attributes with confidence below 0.6 are passed to a larger model, model C (Qwen2.5-VL-72B), for independent judgment and final adjudication~\cite{qwen25vl}. Model C can refer to both models' answers and the discussion history, but is not constrained by their conclusions and may provide a third answer different from both sides. The full automated annotation process is conducted on four RTX 4090 GPUs and takes 7 days to complete.

\begin{table}[!t]
  \caption{Evaluation results on Fine-Grained Object Attribute Understanding across five attribute categories. Struct. Morph., Appear. Surf., Eco. Health, and Pose Inter. denote Structural Morphology, Appearance and Surface, Ecological Health, and Pose and Interaction, respectively. Nereus fine-tuning uses the structured attribute set as the supervision signal. Best values are shown in bold within each backbone.}
    \label{tab:downstream-evaluation}
    \centering
    \scriptsize
    \renewcommand{\arraystretch}{0.88}
    \setlength{\tabcolsep}{2.3pt}
    \resizebox{\linewidth}{!}{%
    \begin{tabular}{lllcccccccccc}
      \toprule
      \multirow{2}{*}{Model} & \multirow{2}{*}{Dataset} & \multirow{2}{*}{Identity} & \multicolumn{2}{c}{Struct. Morph.} & \multicolumn{2}{c}{Appear. Surf.} & \multicolumn{2}{c}{Eco. Health} & \multicolumn{2}{c}{Pose Inter.} & \multicolumn{2}{c}{Overall} \\
      \cmidrule(lr){4-5} \cmidrule(lr){6-7} \cmidrule(lr){8-9} \cmidrule(lr){10-11} \cmidrule(lr){12-13}
      & & Acc $\uparrow$ & Acc $\uparrow$ & Sim $\uparrow$ & Acc $\uparrow$ & Sim $\uparrow$ & Acc $\uparrow$ & Sim $\uparrow$ & Acc $\uparrow$ & Sim $\uparrow$ & Acc $\uparrow$ & Sim $\uparrow$ \\
      \midrule
    BioCLIP2 & Zero-Shot & 22.40 & 6.70 & 19.01 & 5.96 & 10.75 & 12.96 & 13.52 & 3.05 & 23.36 & 10.69 & 22.40 \\
    GPT-5 & Zero-Shot & 18.55 & 20.21 & 53.90 & 31.68 & 50.98 & 60.27 & 81.74 & 25.27 & 49.85 & 30.14 & 59.59 \\
    Gemini 2.0 Flash & Zero-Shot & 13.89 & 19.97 & 56.26 & 25.32 & 48.36 & 13.47 & 69.97 & 18.08 & 49.41 & 18.22 & 58.60 \\
    Gemini 3.1 Pro & Zero-Shot & 32.22 & 28.88 & 61.77 & 34.26 & 55.52 & 64.90 & 88.02 & 33.99 & 53.01 & 37.99 & 65.42 \\
    \midrule
      LLaVA-OneVision-7B & Zero-Shot & 7.07 & 8.16 & 40.81 & 20.06 & 41.32 & 4.97 & 78.41 & 7.84 & 52.08 & 9.28 & 48.39 \\
       & NautData & 14.39 & 8.67 & 44.95 & 22.84 & 47.08 & 4.80 & 67.66 & 6.54 & 47.97 & 11.27 & 50.95 \\
       & MarineInst20M & 23.29 & 10.45 & 45.94 & 33.57 & 53.00 & 1.60 & 22.02 & 19.83 & 30.72 & 15.88 & 42.58 \\
       & \textbf{Nereus (Ours)} & \textbf{63.06} & \textbf{38.30} & \textbf{71.18} & \textbf{54.42} & \textbf{69.52} & \textbf{52.53} & \textbf{83.16} & \textbf{72.77} & \textbf{77.34} & \textbf{51.44} & \textbf{74.80} \\
      \midrule
      InternVL2.5-8B & Zero-Shot & 14.73 & 9.14 & 46.54 & 31.88 & 53.75 & 3.62 & 38.16 & 14.60 & 30.27 & 13.41 & 45.48 \\
       & NautData & 19.38 & 10.17 & 45.51 & 27.90 & 51.29 & 12.21 & 38.39 & 6.54 & 27.21 & 15.13 & 44.36 \\
       & MarineInst20M & 20.80 & 8.58 & 29.92 & 23.14 & 45.27 & 1.35 & 23.02 & 3.70 & 22.85 & 11.67 & 33.98 \\
       & \textbf{Nereus (Ours)} & \textbf{71.80} & \textbf{34.93} & \textbf{70.43} & \textbf{60.68} & \textbf{73.30} & \textbf{48.23} & \textbf{88.53} & \textbf{68.63} & \textbf{76.01} & \textbf{51.88} & \textbf{76.77} \\
      \midrule
      Qwen2.5-VL-7B & Zero-Shot & 17.22 & 9.28 & 43.34 & 29.99 & 45.52 & 6.23 & 70.87 & 18.30 & 45.70 & 14.44 & 51.40 \\
       & NautData & 17.14 & 11.49 & 50.57 & 24.63 & 50.08 & 22.22 & 53.40 & 18.95 & 27.88 & 17.53 & 50.28 \\
       & MarineInst20M & 26.46 & 14.58 & 40.23 & 33.17 & 54.25 & 1.01 & 30.32 & 11.33 & 32.11 & 17.15 & 43.36 \\
       & \textbf{Nereus (Ours)} & \textbf{66.39} & \textbf{30.75} & \textbf{64.30} & \textbf{54.92} & \textbf{67.81} & \textbf{38.47} & \textbf{86.88} & \textbf{65.58} & \textbf{71.43} & \textbf{46.17} & \textbf{73.27} \\
    \bottomrule
    \end{tabular}}
  \end{table}

  \vspace{-0.8em}
  \subsection{Grounded Counting Perception}
  We convert IOCFish5K point-supervised counting data into grounded instruction data for three subtasks: region-aware counting, camouflage-aware counting, and density hotspot reasoning.

  \textbf{{Region-Aware Counting.}} Given fish points, we first use Qwen3-VL-8B-Instruct to identify coarse underwater semantic regions~\cite{qwen3}, including open water, coral shelter, reef or rock structures, seafloor, vegetation, and occluded dark areas. SAM then refines these regions into cleaner masks~\cite{sam}. We assign fish points to the refined regions and to 3$\times$3 grid cells, producing region counts, total counts, densest areas, and spatial comparison statistics for region-aware QA generation.

  \textbf{{Camouflage-Aware Counting.}} We use Qwen3-VL-8B-Instruct to describe region-level visibility rather than semantic categories. Each region is labeled as clear, ambiguous, or highly camouflaged, with reasons such as low contrast, background matching, texture blending, or occlusion. These labels are propagated to fish points, yielding point- and image-level camouflage annotations for visibility-aware counting questions.

  \textbf{{Density Hotspot Reasoning.}} Beyond discrete points, we convert fish points into adaptive Gaussian density maps~{{\cite{mcnn,csrnet}}}, where high-response areas indicate local aggregation. Hotspots are extracted and used as SAM prompts to crop local foreground and background context. Qwen3-VL-8B-Instruct then connects hotspot patches and surrounding scenes with structures such as coral reefs, open water, seafloor, or vegetation. Representative density-hotspot and density-response QA cases are shown in Appendix Figures~\ref{fig:appendix-qa-camouflage-hotspot} and~\ref{fig:appendix-qa-density-reasoning}.

  \textbf{{Grounded Counting Caption Generation.}} We merge RGB images, point annotations, region statistics, camouflage labels, density maps, hotspot information, depth cues, and scene attributions into final annotation records. Qwen3.5-9B generates natural-language QA pairs~\cite{qwen35}, while rule-based templates preserve exact numerical supervision for counts, grid statistics, and visibility summaries. The outputs include annotation JSON files, ShareGPT/LLaVA-style conversations, and VQA datasets, with each answer traceable to evidence such as points, regions, camouflage labels, density hotspots, or local patches.

  \section{Experiment}
  \label{sec:experiment}

  \vspace{-0.8em}
  \subsection{Implementation Details}
  \textbf{Models.}
  We evaluate three open-source VLMs with comparable parameter scales but different architectures: {LLaVA-OneVision-7B~{\cite{llavaov}}, InternVL2.5-8B~{\cite{internvl25}}, and Qwen2.5-VL-7B-Instruct~{\cite{qwen25vl}}}. {For zero-shot baselines, we use their pretrained checkpoints and additionally include BioCLIP2~{\cite{gu2025bioclip2}}, GPT-5~{\cite{openai2025gpt5}}, Gemini 2.0 Flash~{\cite{google2025gemini20flash}}, and Gemini 3.1 Pro~{\cite{google2026gemini31pro}}.}

  \textbf{{Fine-Tuning Strategy.}}
  All comparison and ablation fine-tuning experiments are conducted on four RTX 4090 GPUs and take 12 days in total. We use LoRA while keeping the visual backbone frozen~\cite{lora}; the trainable modules are limited to language-side LoRA layers, with backbone-specific LoRA targets following each model implementation. Additional training settings and hyperparameters are provided in Appendix~\ref{app:training-details}.

\begin{table}[!tb]
    \caption[Nereus counting and density-evidence benchmark.]{{Nereus counting and density-evidence benchmark. Results are grouped by evaluation track, and best values within each track are shown in bold. Acc@{$\pm$}5\% and Acc@{$\pm$}10\% use relative counting-error tolerances of 5\% and 10\%, respectively. For trainable VLMs, QAS combines Track-A counting, spatial, and camouflage supervision, whereas Full retains all QAS components and adds Track-B density, hotspot, and ecological-grounding supervision (Appendix~{\ref{app:qas-full-composition}}). All models use the same Nereus-specific split; evaluation under the official IOCfish5K protocol and the split mapping are provided in Appendix~{\ref{app:official-iocfish-protocol}}.}}
    \label{tab:unified-counting-results}
    \centering
    \scriptsize
    \begin{minipage}[t]{0.59\linewidth}
    \vspace{0pt}
    \centering
    \begingroup
    \setlength{\tabcolsep}{1.2pt}
    \renewcommand{\arraystretch}{0.74}
    \resizebox{\linewidth}{!}{%
    \begin{tabular}{@{}llccccc@{}}
      \toprule
      \rowcolor{black!8}
      \multicolumn{7}{c}{\emph{Track A: RGB counting perception}} \\
      Backbone & Setting & MAE $\downarrow$ & Exact $\uparrow$ & {Acc@{$\pm$}5\%} $\uparrow$ & Cell $\uparrow$ & Camou. F1 $\uparrow$ \\
      \midrule
      CountGD~\cite{countgd2024} & ZS & 32.33 & 6.64 & 14.35 & \textbf{67.67} & --- \\
      CLIP-Count~\cite{clipcount2023} & ZS & 81.20 & 1.06 & 3.90 & 33.22 & --- \\
      \cmidrule(lr){1-7}
      Gemini 2.0 Flash & ZS & 52.80 & 9.08 & 11.12 & 28.79 & 18.94 \\
      & P-ZS & 50.50 & 8.86 & 13.29 & 15.50 & 17.09 \\
      & {Full-Prompt} & 55.50 & 9.04 & 12.68 & 7.62 & 12.73 \\
      \cmidrule(lr){1-7}
      Qwen2.5-VL-7B & ZS & 98.33 & 5.51 & 5.51 & 7.62 & 9.58 \\
      & P-ZS & 94.34 & 7.26 & 7.44 & 4.61 & 2.38 \\
      & Count & 38.57 & 9.92 & 16.83 & 6.11 & 7.41 \\
      & QAS & 31.98 & 10.01 & 18.51 & 49.16 & 52.85 \\
      & Full & 36.19 & 10.36 & 16.56 & 47.39 & \textbf{54.06} \\
      \cmidrule(lr){1-7}
      LLaVA-OneVision-7B & ZS & 113.61 & 4.63 & 4.63 & 0.00 & 10.26 \\
      & P-ZS & 104.38 & 6.24 & 6.24 & 0.18 & 13.33 \\
      & Count & 79.47 & 7.26 & 8.77 & 0.09 & 16.90 \\
      & QAS & 52.81 & 8.06 & 12.84 & 32.68 & 28.91 \\
      & Full & 52.27 & 8.06 & 12.22 & 36.67 & 35.57 \\
      \cmidrule(lr){1-7}
      InternVL2.5-8B & ZS & 87.38 & 2.22 & 2.22 & 22.85 & 21.26 \\
      & P-ZS & 167.55 & 0.00 & 1.16 & 20.73 & 18.98 \\
      & Count & 31.72 & 10.01 & 19.04 & 20.81 & 14.72 \\
      & QAS & \textbf{25.99} & \textbf{10.45} & \textbf{20.46} & {52.70} & 50.02 \\
      & Full & 30.48 & 9.74 & 18.95 & 51.90 & 52.59 \\
      \bottomrule
    \end{tabular}}
    \endgroup
    \end{minipage}\hfill
    \begin{minipage}[t]{0.38\linewidth}
    \vspace{0pt}
    \centering
    \begingroup
    \setlength{\tabcolsep}{1.0pt}
    \renewcommand{\arraystretch}{1.0347}
    \resizebox{\linewidth}{!}{%
    \begin{tabular}{@{}llccc@{}}
      \toprule
      \rowcolor{black!8}
      \multicolumn{5}{c}{\emph{Track B: RGB+heatmap density evidence}} \\
      Backbone & Setting & Hotspot $\uparrow$ & Density F1 $\uparrow$ & Ground. F1 $\uparrow$ \\
      \midrule
      Gemini 2.0 Flash & ZS & 44.88 & 18.35 & 16.48 \\
      & {Full-Prompt} & 41.91 & 18.22 & 16.15 \\
      \cmidrule(lr){1-5}
      Qwen2.5-VL-7B & ZS & 41.75 & 18.13 & 16.04 \\
      & D-QA & 37.39 & \textbf{45.26} & \textbf{37.19} \\
      & Full & 35.37 & 45.24 & 37.14 \\
      \cmidrule(lr){1-5}
      LLaVA-OneVision-7B & ZS & 32.86 & 18.98 & 17.34 \\
      & D-QA & 1.42 & 28.27 & 23.90 \\
      & Full & \textbf{47.48} & 30.32 & 25.22 \\
      \cmidrule(lr){1-5}
      InternVL2.5-8B & ZS & 42.13 & 19.80 & 18.37 \\
      & D-QA & 37.59 & \textbf{45.26} & 36.35 \\
      & Full & 38.70 & 43.24 & 35.00 \\
      \bottomrule
    \end{tabular}}
    \endgroup

    \vspace{1.6pt}

    \begingroup
    \setlength{\tabcolsep}{1.0pt}
    \renewcommand{\arraystretch}{0.88}
    \resizebox{\linewidth}{!}{%
    \begin{tabular}{@{}lccccc@{}}
      \toprule
      \rowcolor{black!8}
      \multicolumn{6}{c}{\emph{Track C: Nautilus transfer}} \\
      Setting & MAE $\downarrow$ & Exact $\uparrow$ & {Acc@{$\pm$}5\%} $\uparrow$ & Cell $\uparrow$ & Camou. F1 $\uparrow$ \\
      \midrule
      \shortstack[l]{Nautilus\\Counting SFT} & \raisebox{0.55ex}{39.96} & \raisebox{0.55ex}{9.65} & \raisebox{0.55ex}{16.30} & \raisebox{0.55ex}{6.11} & \raisebox{0.55ex}{8.48} \\
      \cmidrule(lr){1-6}
      \addlinespace[1pt]
      Nereus Count & 38.57 & 9.92 & 16.83 & 6.11 & 7.41 \\[1.5pt]
      Nereus QAS & \textbf{31.98} & 10.01 & \textbf{18.51} & \textbf{49.16} & 52.85 \\[1.5pt]
      Nereus Full & 36.19 & \textbf{10.36} & 16.56 & 47.39 & \textbf{54.06} \\[1.5pt]
      \bottomrule
    \end{tabular}}
    \endgroup
    \end{minipage}
    \vspace{2pt}
    \parbox{0.97\linewidth}{\scriptsize
    ZS: zero-shot; P-ZS: prompted zero-shot; Count: count-only SFT; QAS: QA+structured-final SFT; D-QA: density QA SFT; {Full: full Nereus SFT for trainable VLMs; Full-Prompt: Gemini API inference with the complete Nereus task protocol and structured output requirements, without parameter fine-tuning.} {---: not applicable.}}
  \end{table}

  \textbf{{Evaluation Protocols.}}
  We evaluate Nereus on two QA-based task families: Fine-Grained Object Attribute Understanding and Grounded Counting Perception, both designed for downstream marine applications. Each sample contains an underwater visual input, optional task-specific grounding evidence, and a natural-language question. Outputs are scored with task-specific normalization, synonym mapping, structured parsing, and LLM-as-a-judge~{{\cite{llmjudge}}} for free-form ecological or grounding responses. For attribute understanding, models predict object-level biological attributes from the image and target localization across five aspects: identity, structural morphology, appearance and surface, ecological health, and pose and interaction. We report category-level accuracy, overall accuracy, and semantic similarity. {For counting perception, we evaluate region-aware counting, camouflage-state counting, and density-hotspot reasoning using Count MAE, Exact Count Accuracy, Acc@{$\pm$}5\%, Acc@{$\pm$}10\%, Densest Cell Accuracy, Camouflage F1, Density F1, and Ecology/Grounding F1. Acc@{$\pm$}5\% and Acc@{$\pm$}10\% measure the proportion of predictions within 5\% and 10\% of the ground-truth count, respectively.} Track A uses RGB images to test numerical counting, densest-cell localization, and camouflage-aware reasoning under zero-shot, prompted zero-shot, count-only SFT, QA+structured-final SFT, and full Nereus SFT settings. Track B adds density heatmaps for hotspot localization, density interpretation, and ecological grounding under density zero-shot, density-QA SFT, and full Nereus SFT settings. We also fine-tune Qwen2.5-VL-7B on Nautilus counting QA to evaluate external transfer. The evaluated models include {LLaVA-OneVision-7B, InternVL2.5-8B, Qwen2.5-VL-7B, and Gemini 2.0 Flash.}

  \subsection{Main Results}

  \textbf{{Fine-Grained Object Attribute Understanding.}} {Table~{\ref{tab:downstream-evaluation}} compares Nereus with NautData and MarineInst20M across five attribute categories. We align the training sample scale across datasets and fine-tune LLaVA-OneVision-7B, InternVL2.5-8B, and Qwen2.5-VL-7B, alongside additional zero-shot baselines. Across the three open-source backbones, Nereus improves overall accuracy and semantic similarity over zero-shot evaluation and adaptation with the two existing datasets. These gains support the value of structured biological supervision for Underwater Fine-grained Perception. Structural morphology remains challenging because it depends on subtle visual cues, such as fin shape, mouth type, and coral growth form, which are easily affected by underwater degradation.}

  \textbf{{Grounded Counting Perception.}} {Table~{\ref{tab:unified-counting-results}} reports counting, spatial localization, camouflage reasoning, and density-evidence results. On Track A, Nereus QAS reduces InternVL2.5-8B's MAE from 87.38 to 25.99, outperforming CountGD's MAE of 32.33. CountGD remains particularly strong in densest-cell localization, highlighting the complementary strengths of specialized counting models and Nereus-trained VLMs. On Track B, density QA supervision improves Qwen2.5-VL-7B's Density F1 from 18.13 to 45.26 and Grounding F1 from 16.04 to 37.19. In the Nautilus transfer comparison, Nereus QAS achieves lower MAE than Nautilus counting SFT (31.98 versus 39.96) and higher densest-cell accuracy (49.16\% versus 6.11\%). These results demonstrate the benefits of combining numerical counting with spatial and ecological supervision.}

  \subsection{Ablation Study}
  We further study how different supervision formats and degradation-aware inputs affect model performance. For Fine-Grained Object Attribute Understanding, we fine-tune Qwen2.5-VL-7B with three supervision formats at a comparable data scale: single-attribute QA pairs, object-level captions, and structured attribute sets. Table~\ref{tab:supervisory-signal} compares these supervision signals under strict accuracy and semantic similarity.
  
  \begin{table}[!htbp]
    \caption{Ablation study on different supervisory signals for Fine-Grained Object Attribute Understanding using Qwen2.5-VL-7B. Object-level Caption denotes natural-language object descriptions. Acc is averaged over all attribute questions within each category.}
    \label{tab:supervisory-signal}
    \centering
    \scriptsize
    \setlength{\tabcolsep}{2.8pt}
    \resizebox{\linewidth}{!}{%
    \begin{tabular}{lccccccccccc}
      \toprule
      \multirow{2}{*}{Supervisory Signal} & \multirow{2}{*}{Identity} & \multicolumn{2}{c}{Struct. Morph.} & \multicolumn{2}{c}{Appear. Surf.} & \multicolumn{2}{c}{Eco. Health} & \multicolumn{2}{c}{Pose Inter.} & \multicolumn{2}{c}{Overall} \\
      \cmidrule(lr){3-4} \cmidrule(lr){5-6} \cmidrule(lr){7-8} \cmidrule(lr){9-10} \cmidrule(lr){11-12}
      & Acc $\uparrow$ & Acc $\uparrow$ & Sim $\uparrow$ & Acc $\uparrow$ & Sim $\uparrow$ & Acc $\uparrow$ & Sim $\uparrow$ & Acc $\uparrow$ & Sim $\uparrow$ & Acc $\uparrow$ & Sim $\uparrow$ \\
      \midrule
      QA Pairs & 10.98 & \textbf{33.64} & 49.80 & 33.38 & 52.00 & \textbf{77.90} & 82.00 & \textbf{52.13} & 70.80 & 39.08 & 55.20 \\
      Object-level Caption & 20.99 & 17.18 & 57.70 & 27.87 & 55.50 & 3.94 & 64.60 & 11.93 & 68.00 & 16.78 & 61.60 \\
      Structured Attribute Set & \textbf{58.06} & 30.84 & \textbf{64.00} & \textbf{50.00} & \textbf{67.10} & 39.92 & \textbf{87.60} & 34.41 & \textbf{72.10} & \textbf{41.55} & \textbf{73.10} \\
      \bottomrule
    \end{tabular}}
  \end{table}

  The results show a useful distinction between exact matching and semantic consistency. QA-pair supervision obtains strong accuracy on ecological-health and pose-interaction attributes, likely because these categories contain more repeated answer patterns and the QA format directly matches the evaluation query. However, its identity accuracy is low, since species prediction has a much larger and more long-tailed label space, and isolated QA pairs provide weaker cross-attribute constraints. In contrast, structured attribute supervision is not the best on every strict-accuracy column, but it achieves the best overall accuracy and consistently higher semantic similarity. This suggests that structured annotations help the model learn relations between visual evidence, species identity, morphology, appearance, and interaction attributes, so even when an exact normalized label is missed, the prediction remains closer in meaning.

\begin{figure}[!tb]
    \centering
    \includegraphics[width=0.42\linewidth]{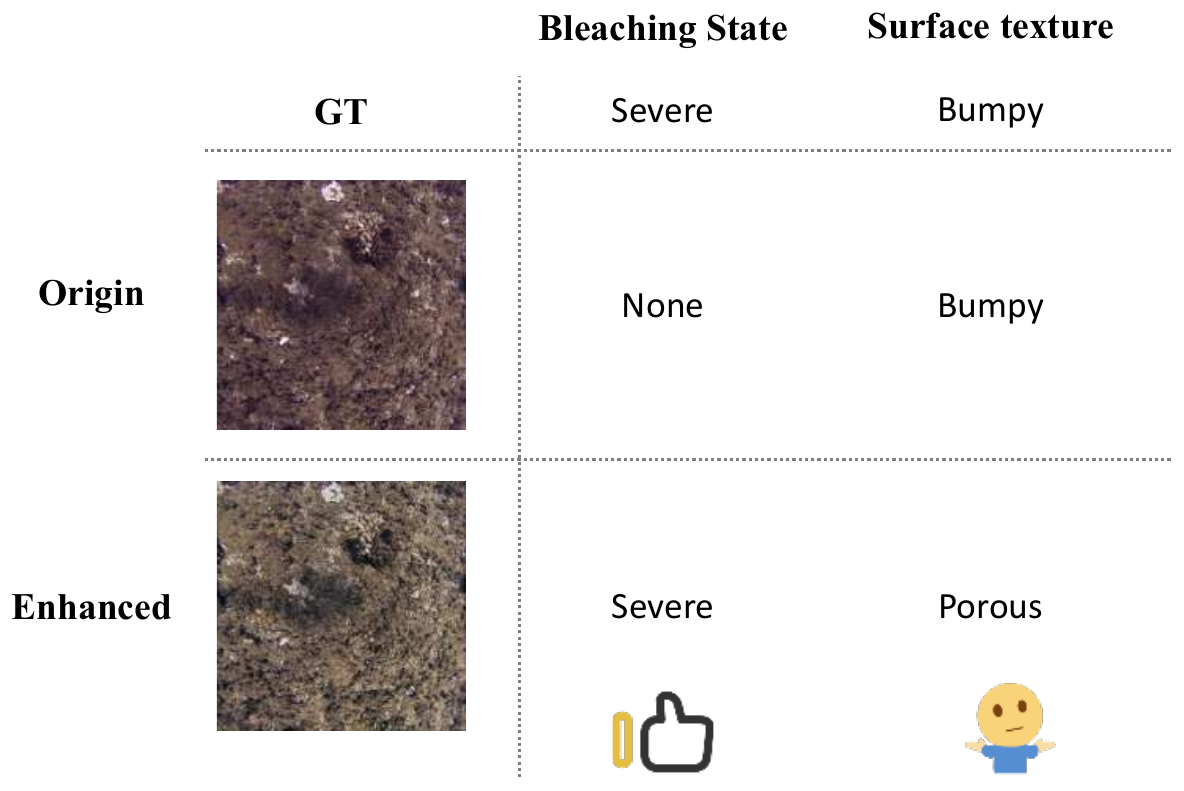}
    \caption{Degradation examples.}
    \label{fig:degradation-examples}
  \end{figure}

  Figure~\ref{fig:degradation-examples} shows representative underwater degradation examples in Nereus. The figure remains in the main text to illustrate the visual conditions considered by Nereus, while the detailed degradation-aware setup, full results, and interpretation are provided in Appendix~\ref{app:degradation-ablation-results}.

  For Grounded Counting Perception, we conduct an additional image-only ablation on InternVL2.5-8B to isolate the effect of different counting supervision forms. As shown in Appendix Table~\ref{tab:internvl-image-only-ablation}, count-only supervision mainly improves numerical counting, reducing MAE from 87.38 to 31.72, but brings limited gains in spatial and camouflage reasoning. QA+structured-final supervision further lowers MAE to 25.99 and greatly improves Densest Cell Accuracy and Camouflage F1, showing that region, grid, and visibility annotations help the model learn grounded counting evidence rather than only total counts. The full Nereus setting achieves the best Camouflage F1, indicating that richer supervision is particularly useful for visibility-aware reasoning.

  \section[Limitations]{\texorpdfstring{{Limitations}}{Limitations}}
  \label{sec:limitations}
  {Nereus may underrepresent taxa, regions, depths, and imaging devices; robustness to rare species and atypical individuals, including injured organisms, remains insufficiently evaluated. Random train/test sampling does not explicitly separate locations, dates, or dives. Source-specific filtering improves annotation reliability but may bias coverage relative to unfiltered imagery.}

  {Automatic annotations may be noisy, and image-level degradation indicators cannot fully capture local water quality or hidden biological states. Expert review covers only a sampled subset due to time and expertise constraints; remaining annotation errors are unquantified. Nereus primarily supports fine-tuning rather than training LMMs from scratch. Future work will expand expert verification and evaluate unfiltered continuous video and broader real-world deployments.}

  \section{Conclusion}

  We introduce Nereus, a large-scale underwater dataset for Fine-Grained Object Attribute Understanding and Grounded Counting Perception. Nereus contains 88K images and 2.68M {question-answer pairs}, covering 136K marine organism objects across 2,690 species and 1,515 genera, as well as 65K point annotations for grounded counting. Built around downstream marine applications, Nereus provides fine-grained biological attributes, region- and camouflage-aware counting supervision, density reasoning evidence, and image-level degradation labels. We further benchmark representative open-source and {proprietary LMMs}, showing that training on Nereus improves {Underwater Fine-grained Perception} while current models still struggle with subtle attributes, grounding, degradation, and reasoning.

  More broadly, Nereus connects organism-level biological understanding with spatial evidence and underwater-specific visual conditions, offering a unified resource for studying marine perception beyond generic captioning or total-count prediction. We expect it to support future work on reliable ecological monitoring, degradation-aware model adaptation, and multimodal reasoning over dense underwater scenes. Its annotations and protocols also support model diagnosis and expert-guided marine analysis. Limitations and future directions are discussed in Section~\ref{sec:limitations}.

  \renewcommand{\acksection}{\section*{\texorpdfstring{{Acknowledgments and Disclosure of Funding}}{Acknowledgments and Disclosure of Funding}}}
  \begin{ack}
    {This work was partially supported by the National Natural Science Foundation of China (Grant No. 62206123) and the Key Science and Technology Program of Lhasa Municipality (Grant No. LSKJ202612).}
  \end{ack}

  \clearpage
  \bibliographystyle{unsrt_camera_ready}
  \bibliography{reference_camera_ready,reference_camera_ready_additions}

  \clearpage
  \appendix

  \begin{center}
    {\LARGE\bfseries Nereus\,\raisebox{-0.12em}{\includegraphics[height=0.95em]{photos/camera_ready/jellyfish_title.png}}: A Large-Scale Underwater Dataset for}\\[-0.1em]
    {\LARGE\bfseries Fine-Grained Attribute Understanding and}\\[-0.1em]
    {\LARGE\bfseries Grounded Counting Perception}\\[0.35em]
    {\Large\bfseries (Supplementary material)}
  \end{center}

  \vspace{2.0em}
  {\Large\bfseries Table of Contents in Appendix}
  \vspace{1.6em}

  \appendixcontentssection{A}{Dataset, Training, and Release Details}{app:dataset-training-release}
  \appendixcontentssubsection{A.1}{{Evaluation Scales for Annotation Quality}}{app:evaluation-scales}
  \appendixcontentssubsection{A.2}{{Training Details}}{app:training-details}
  \appendixcontentssubsection{A.3}{{License Details}}{app:license-details}
  \appendixcontentssubsection{A.4}{{Broader Impacts}}{app:broader-impacts}
  \appendixcontentssubsection{A.5}{{Additional Data Processing Details}}{app:data-processing-details}
  \appendixcontentssubsection{A.6}{{Knowledge-Prior Prompt Example}}{app:kg-prior-example}
  \appendixcontentssubsection{A.7}{{Degradation Indicator Details}}{app:degradation-details}
  \appendixcontentssubsection{A.8}{{Degradation-Aware Ablation Results}}{app:degradation-ablation-results}
  \appendixcontentssubsection{A.9}{{Expert Review and Inter-Annotator Agreement}}{app:expert-review-iaa}

  \vspace{0.6em}
  \appendixcontentssection{B}{Counting and Density-Evidence Benchmark}{app:counting-benchmark}
  \appendixcontentssubsection{B.1}{{Expanded Nautilus Transfer Comparison}}{app:expanded-nautilus-transfer}
  \appendixcontentssubsection{B.2}{{Specialized Counting Baselines}}{app:specialized-counting-baselines}
  \appendixcontentssubsection{B.3}{{Result Interpretation}}{app:result-interpretation}
  \appendixcontentssubsection{B.4}{{Qualitative QA Examples}}{app:qualitative-qa-examples}
  \appendixcontentssubsection{B.5}{{Metric Definitions}}{app:metric-definitions}
  \appendixcontentssubsection{B.6}{{Task Composition of QAS and Full}}{app:qas-full-composition}

  \vspace{0.6em}
  \appendixcontentssection{C}{{Data Splits and Additional Evaluation}}{app:split-additional-evaluation}
  \appendixcontentssubsection{C.1}{{Official IOCfish5K Protocol}}{app:official-iocfish-protocol}

  \vspace{0.6em}
  \appendixcontentssection{D}{NeurIPS Paper Checklist}{app:checklist}

  \clearpage
  \section{Dataset, Training, and Release Details}
  \label{app:dataset-training-release}

  \begin{figure}[H]
    \centering
    \includegraphics[width=0.78\linewidth]{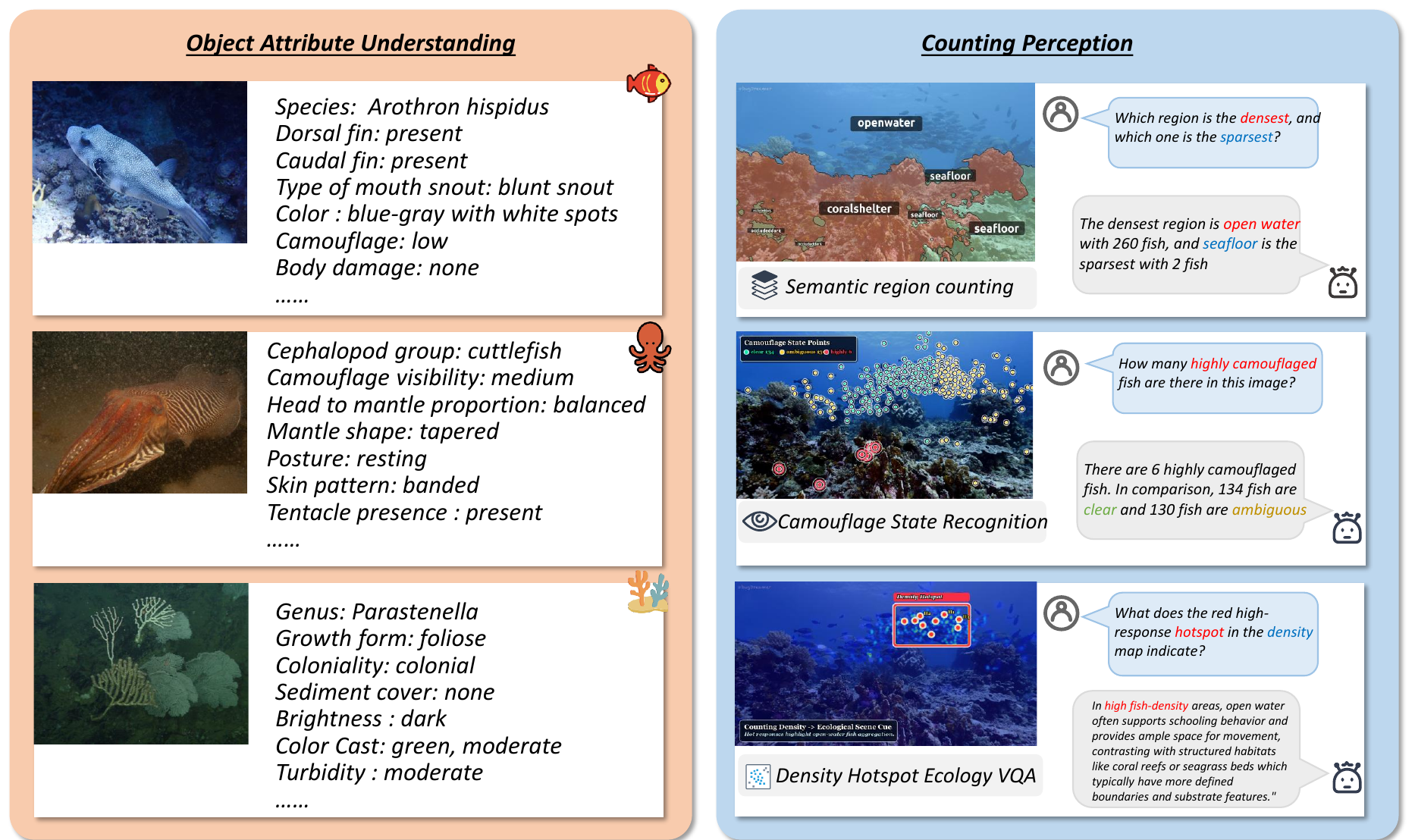}
    \caption{Examples of Nereus applications, covering object-level attribute understanding, region-aware counting, camouflage-aware counting, density-hotspot reasoning, and underwater degradation annotation.}
    \label{fig:task-display}
  \end{figure}

  \begin{figure}[H]
    \centering
    \includegraphics[width=0.95\linewidth]{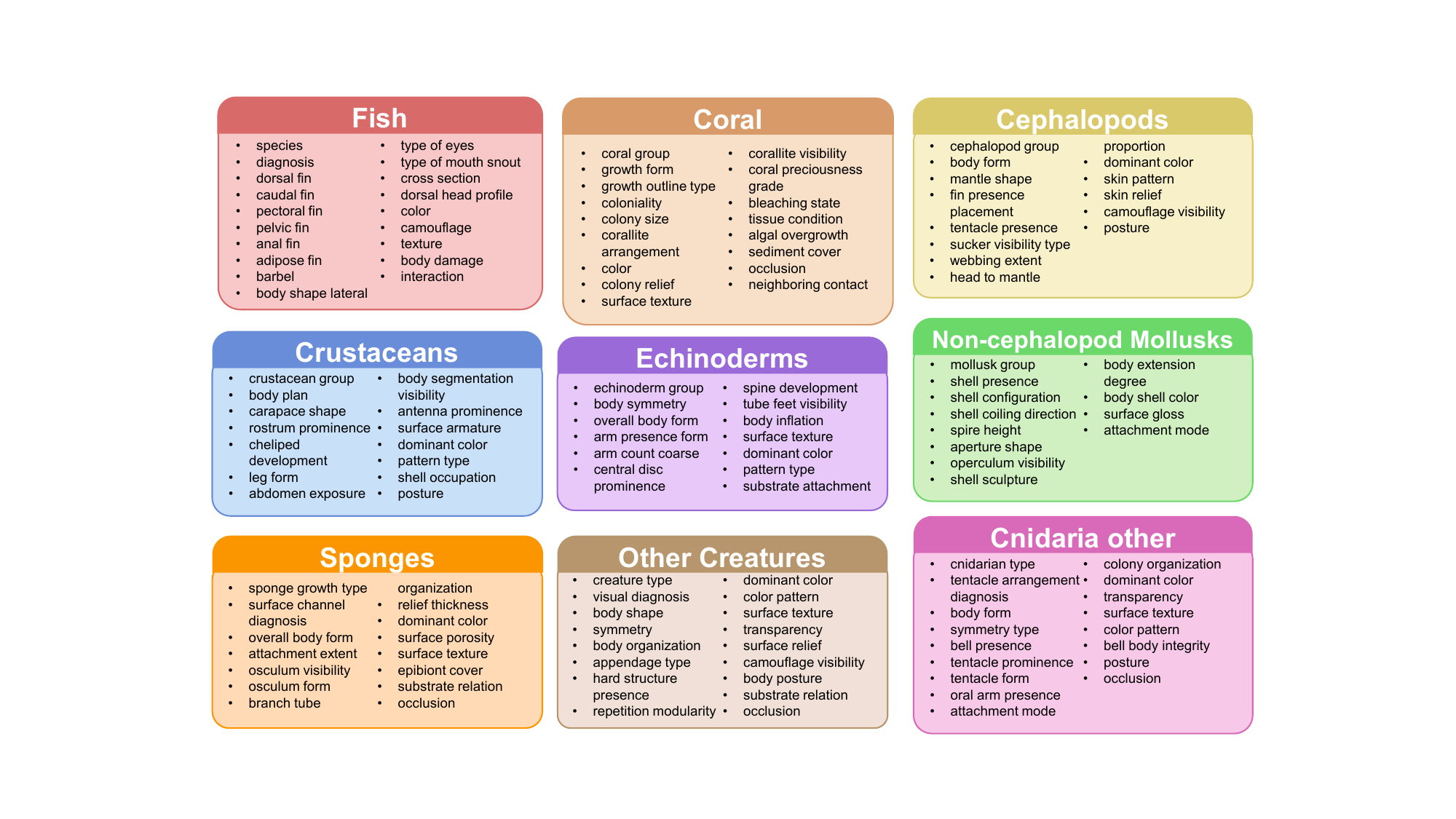}
    \caption{Overview of the Nereus attribute taxonomy. The taxonomy organizes the 136-dimensional structured attribute set into identity, structural morphology, appearance and surface, condition and environment, and pose, visibility, and interaction categories.}
    \label{fig:attribute-taxonomy}
  \end{figure}

  \subsection[Evaluation Scales for Annotation Quality]{\texorpdfstring{{Evaluation Scales for Annotation Quality}}{Evaluation Scales for Annotation Quality}}
  \label{app:evaluation-scales}
  We use the following calibrated scales to evaluate annotation quality, support human verification, and compute semantic similarity for free-form underwater attribute responses.

  \begin{table}[H]
    \centering
    \small
    \renewcommand{\arraystretch}{1.28}
    \caption{10-point scale for automated annotation quality assessment.}
    \label{tab:10-point-scale}
    \begin{tabularx}{\textwidth}{c X}
      \toprule
      \textbf{Score} & \textbf{Criterion and descriptive requirement} \\
      \midrule
      10 & Perfectly accurate, complete, internally consistent, and highly aligned with image evidence. \\
      9  & Accurate overall, with only trivial imperfections that do not affect the core understanding. \\
      8  & Mostly accurate; minor issues, such as slight terminology bias, but no critical errors. \\
      7  & Mostly correct; some secondary attributes show deviations, but core understanding is intact. \\
      6  & Partially correct; the core category is right, but notable errors or omissions affect fine-grained details. \\
      5  & Borderline usable; only basic attributes are trustworthy, with frequent deviations. \\
      4  & Poor; multiple obvious errors, substantial omissions, and few correct attributes remain. \\
      3  & Bad; major attributes are wrong and misleading overall, with only incidental correct statements. \\
      2  & Very bad; the vast majority of data is wrong, missing, or contradictory. \\
      1  & Completely wrong or irrelevant; contradicts image evidence with no redeeming accuracy. \\
      \bottomrule
    \end{tabularx}
  \end{table}

  \begin{table}[H]
    \centering
    \small
    \renewcommand{\arraystretch}{1.28}
    \caption{Four-level scale for manual annotation accuracy and mapping.}
    \label{tab:4-level-scale}
    \begin{tabularx}{\textwidth}{c c X}
      \toprule
      \textbf{Grade} & \textbf{Mapping score} & \textbf{Qualitative criterion} \\
      \midrule
      \textbf{A} & 10 & Accurate: reflects visual features well, is consistent with the image, and has only minor flaws. \\
      \textbf{B} & 7  & Acceptable: generally correct, captures main features, and supports basic understanding. \\
      \textbf{C} & 4  & Deficient: partially correct but contains obvious errors and has limited reference value. \\
      \textbf{D} & 1  & Failed: inconsistent with the image, with severe errors or internal conflicts. \\
      \bottomrule
    \end{tabularx}
  \end{table}

  \begin{table}[H]
    \centering
    \small
    \renewcommand{\arraystretch}{1.28}
    \caption{100-point scale for SIM-based semantic similarity assessment.}
    \label{tab:100-point-sim-scale}
    \begin{tabularx}{\textwidth}{c X}
      \toprule
      \textbf{Score range} & \textbf{Semantic alignment criterion} \\
      \midrule
      90--100 & Semantically equivalent, synonymous, or fully correct. \\
      70--89  & Correct core meaning with minor differences in granularity or phrasing. \\
      50--69  & Correct direction or intent but imprecise or slightly vague. \\
      30--49  & Related to the topic but factually or logically clearly wrong. \\
      10--29  & Nearly irrelevant; contains only marginal keywords related to the domain. \\
      0--9    & Completely wrong, empty, or nonsensical. \\
      \bottomrule
    \end{tabularx}
  \end{table}

  \subsection[Training Details]{\texorpdfstring{{Training Details}}{Training Details}}
  \label{app:training-details}
  We use LoRA-based supervised fine-tuning for all open-source backbones~\cite{lora}. The visual backbone remains frozen across models. For Qwen2.5-VL-7B, training is conducted with LLaMA-Factory~\cite{llamafactory}, and LoRA adapters are applied to language-side attention and MLP layers. For InternVL2.5-8B, we use the native InternVL training framework and train only LLM LoRA parameters, while freezing the LLM base weights, the cross-modal MLP projector, and the vision backbone. For LLaVA-OneVision-7B, we use HuggingFace Trainer with PEFT~\cite{peft}, freeze the vision tower and multimodal projector, and train language-model LoRA parameters. The exact learning rates, LoRA ranks, batch sizes, sequence lengths, and image-resolution constraints follow the corresponding backbone and task settings. Across experiments, we use BF16 precision and train on RTX 4090 GPUs.

  \subsection[License Details]{\texorpdfstring{{License Details}}{License Details}}
  \label{app:license-details}
  {FathomNet images and source annotations retain their applicable original licenses}~\cite{fathomnet}. Following these license requirements, we do not directly package or redistribute the original FathomNet images in our dataset release. Instead, we retain the FathomNet image URLs in the annotation files so that users can download the images themselves. CoralVQA~{{\cite{coralvqa}}} and iNaturalist~{{\cite{inaturalist}}} data follow their respective open licenses. {IOCfish5K images and source annotations retain their CC BY-NC 4.0 license. Nereus-authored attribute annotations and counting QA data are released under CC BY-NC 4.0 to the extent of our rights. This license does not replace or broaden the licenses governing third-party images and source annotations.}

  \subsection[Broader Impacts]{\texorpdfstring{{Broader Impacts}}{Broader Impacts}}
  \label{app:broader-impacts}
  Nereus is intended to support research on underwater visual understanding, marine biodiversity analysis, ecological conservation, fishery monitoring, and science communication. By providing fine-grained organism attributes, grounded counting evidence, and degradation-aware annotations, the dataset may reduce the cost of manual visual inspection and help researchers study abundance, habitat association, visibility conditions, and organism-level traits at larger scale. These capabilities can benefit ecological assessment, long-term reef and fishery monitoring, educational tools, and assistive systems for marine scientists.

  At the same time, models trained on Nereus should be used with care. Fine-grained recognition and counting systems may be unreliable under domain shift, rare taxa, severe turbidity, unusual imaging devices, or regions underrepresented in the source data. Incorrect predictions could mislead ecological interpretation if used without expert review. The dataset may also be misused for intensive resource extraction, intrusive monitoring of sensitive habitats, or disclosure of locations associated with vulnerable species when combined with external metadata. In addition, LMM-assisted annotations and automatic preprocessing can introduce systematic errors, and image-level degradation indicators cannot fully describe local water quality or hidden biological states.

  We therefore recommend using Nereus as a research and decision-support resource rather than a replacement for expert ecological assessment. Users should respect the licenses of original image sources, audit model behavior before deployment, report uncertainty in high-stakes settings, and avoid releasing sensitive location information for vulnerable species or habitats. Future releases should continue expanding taxonomic, geographic, depth, and device coverage, and should include broader expert validation where the dataset is used for conservation or policy-relevant decisions.

  \subsection[Additional Data Processing Details]{\texorpdfstring{{Additional Data Processing Details}}{Additional Data Processing Details}}
  \label{app:data-processing-details}
  {Preprocessing differs across sources according to the available localization annotations.}

  For CoralVQA, which only provides point annotations, bounding boxes are unavailable. We therefore retain only images with a single genus label, meaning that all point annotations in the same image belong to the same genus. This ensures that the whole image can serve as a valid annotation unit for that genus. During data cleaning, we correct label spelling errors, such as correcting Porties to Porites and Orcbicella to Orbicella. We also split labels with bleaching prefixes, such as ``Bleached Montipora,'' into two independent fields: genus and bleaching status.

  {For sources with existing object localization, we use the original annotations or manually reviewed metadata. For iNaturalist images, which lack ground-truth bounding boxes and point annotations, we use GroundingDINO to detect target objects and obtain bounding boxes.} We then apply the same patch cropping and quality filtering pipeline as used for FathomNet.

  \subsection[Knowledge-Prior Prompt Example]{\texorpdfstring{{Knowledge-Prior Prompt Example}}{Knowledge-Prior Prompt Example}}
  \label{app:kg-prior-example}
  During annotation, the concept name of each object is used to retrieve the corresponding knowledge-graph entry, which is then inserted into the model prompt as reference context. The following example shows the prior prompt for \emph{Merluccius productus} and the resulting structured annotation fields.

  \noindent\textbf{{KG Prior Prompt.}}
  \begin{kgpriorbox}
  \small Knowledge prior for \emph{Merluccius productus} (Fish): habitat: pelagic-neritic; max length: 83.0 cm; family: Merlucciidae; order: Gadiformes. Note: these are reference values; actual observation may differ.
  \end{kgpriorbox}

  \noindent\textbf{{Final Annotations.}}
  \begin{center}
  \small
  \begin{tabular}{@{}ll@{}}
    \toprule
    Field & Value \\
    \midrule
    adipose\_fin & absent \\
    anal\_fin & absent \\
    barbel & absent \\
    body\_damage & none \\
    body\_shape\_lateral & elongate \\
    camouflage & low \\
    caudal\_fin & present \\
    color & silver-gray \\
    cross\_section & circular \\
    diagnosis & Silver-gray body with black speckling \\
    dorsal\_fin & present \\
    dorsal\_head\_profile & straight \\
    interaction & [schooling] \\
    pectoral\_fin & absent \\
    pelvic\_fin & absent \\
    species & \emph{Merluccius productus} \\
    texture & smooth \\
    type\_of\_eyes & small-eyed \\
    type\_of\_mouth\_snout & terminal mouth \\
    \bottomrule
  \end{tabular}
  \end{center}

  \subsection[Degradation Indicator Details]{\texorpdfstring{{Degradation Indicator Details}}{Degradation Indicator Details}}
  \label{app:degradation-details}
  To support evaluation experiments on degradation-aware perception, we compute image-level degradation indicators for all {88K images} in the dataset. Following the underwater image quality assessment method introduced by UIQI, we extract quality-aware features along three degradation dimensions: brightness, color cast, and turbidity~\cite{uiqi}.

  \textbf{Brightness.}
  We first convert each image to grayscale and compute its normalized mean luminance:
  \[
    l = \frac{1}{|\Omega|}\sum_{x \in \Omega}\frac{I_{\mathrm{gray}}(x)}{255}, \quad l \in [0,1],
  \]
  where a larger value indicates a brighter image.

  \textbf{{Color Cast.}}
  We convert the image to the CIELAB color space and compute the pixel-wise means $m_a, m_b$ and variances $v_a, v_b$ of the $a$ and $b$ channels. The overall color mean $m$, color variance $v$, and color-cast degree $d$ are defined as
  \[
    m = \sqrt{m_a^2 + m_b^2}, \quad
    v = \sqrt{v_a^2 + v_b^2}, \quad
    d = \frac{m}{v + \epsilon}.
  \]
  A larger $d$ indicates a more severe color cast. The color direction is determined by the polar angle on the $a$-$b$ chromaticity plane,
  \[
    \theta = \operatorname{atan2}(m_b, m_a),
  \]
  which is used to categorize the color cast into blue, green, blue-green, yellow, and other types.

  \textbf{Turbidity.}
  Underwater scattering often causes haze, contrast reduction, and loss of fine details. We capture these degradation patterns with five complementary features: (1) the high-frequency sub-band energy $E_T$ obtained from three-level CDF 9/7 wavelet decomposition, which reflects local blur; (2--3) the shape parameter $\lambda$ and scale parameter $k$ obtained by fitting a generalized Gaussian distribution to mean-subtracted contrast-normalized coefficients, which reflect global haze; (4) the local contrast energy $CE$ computed on the gray, yb, and rg channels; and (5) the Jensen-Shannon divergence $D_{JS}$ between the grayscale histogram and the uniform distribution, which reflects global contrast.

  To avoid interference from brightness and color cast when estimating turbidity, all five features are computed on preprocessed images. The preprocessing includes brightness normalization, which rescales the mean luminance to 128, and gray-world-based white balance. The standardized features are then fused by principal component analysis (PCA), and the first principal component is used as the final turbidity score:
  \[
    s_{\mathrm{tur}} = \operatorname{PC}_1\left(
    \operatorname{standardize}\left([E_T, \lambda, k, CE, D_{JS}]^\top\right)
    \right).
  \]

  \subsection[Degradation-Aware Ablation Results]{\texorpdfstring{{Degradation-Aware Ablation Results}}{Degradation-Aware Ablation Results}}
  \label{app:degradation-ablation-results}

  We investigate whether degradation priors and image enhancement can help Fine-Grained Object Attribute Understanding. We compare the original-image baseline with two degradation-aware settings: adding degradation type and level as textual priors, and applying DTIUIE enhancement together with the same priors~\cite{dtiuie}. Table~\ref{tab:appendix-degradation-ablation} reports the degradation-aware training ablation for Ultra Fine-Grained Object Attribute Understanding using Qwen2.5-VL-7B, with representative degradation examples shown in Figure~\ref{fig:degradation-examples}.

  \begin{table}[H]
    \caption{Ablation study on degradation-aware training strategies for Ultra Fine-Grained Object Attribute Understanding using Qwen2.5-VL-7B.}
    \label{tab:appendix-degradation-ablation}
    \centering
    \scriptsize
    \setlength{\tabcolsep}{2.8pt}
    \resizebox{\linewidth}{!}{%
    \begin{tabular}{lccccccccccc}
      \toprule
      \multirow{2}{*}{Setting} & \multirow{2}{*}{Identity} & \multicolumn{2}{c}{Struct. Morph.} & \multicolumn{2}{c}{Appear. Surf.} & \multicolumn{2}{c}{Eco. Health} & \multicolumn{2}{c}{Pose Inter.} & \multicolumn{2}{c}{Overall} \\
      \cmidrule(lr){3-4} \cmidrule(lr){5-6} \cmidrule(lr){7-8} \cmidrule(lr){9-10} \cmidrule(lr){11-12}
      & Acc $\uparrow$ & Acc $\uparrow$ & Sim $\uparrow$ & Acc $\uparrow$ & Sim $\uparrow$ & Acc $\uparrow$ & Sim $\uparrow$ & Acc $\uparrow$ & Sim $\uparrow$ & Acc $\uparrow$ & Sim $\uparrow$ \\
      \midrule
      Original Img. & \textbf{58.06} & \textbf{30.84} & 64.00 & \textbf{50.00} & \textbf{67.10} & 39.92 & 87.60 & 34.41 & 72.10 & \textbf{41.55} & 73.10 \\
      Degradation Prior + Original Img. & 56.30 & 29.30 & \textbf{66.20} & 48.16 & 66.80 & \textbf{40.25} & \textbf{89.50} & 35.95 & 76.20 & 40.51 & \textbf{74.60} \\
      Degradation Prior + Enhanced Img. & 51.95 & 25.27 & 62.30 & 45.56 & 66.20 & 38.18 & 86.40 & \textbf{36.12} & \textbf{76.40} & 37.36 & 71.80 \\
      \bottomrule
    \end{tabular}}
  \end{table}

  Degradation priors improve semantic similarity across most attribute categories, with clear gains on structural morphology, ecological health, and pose-interaction attributes. This suggests that explicit degradation information can help the model interpret ambiguous visual cues affected by turbidity, color cast, or blur. However, using enhanced images does not further improve performance and instead slightly degrades most accuracy and similarity scores. This indicates that image enhancement for downstream marine understanding should be task-driven rather than optimized only for perceptual visual quality.

  \Needspace{18\baselineskip}
  \subsubsection[Exploratory Degradation Probes]{\texorpdfstring{{Exploratory Degradation Probes}}{Exploratory Degradation Probes}}
  \label{app:degradation-probes}
  \par
  As an exploratory study to inform future work on degradation-aware perception, we evaluate linear and KNN probes on 1,000 evaluation objects using UIQI-derived low-level visual-quality features~\cite{uiqi}. Table~\ref{tab:degradation-probes} reports accuracy and balanced accuracy for turbidity and color-cast classification.
  \par

  \begin{table}[H]
    \caption[Exploratory degradation classification using UIQI-derived features.]{{Exploratory degradation classification using UIQI-derived features. All values are percentages.}}
    \label{tab:degradation-probes}
    \centering
    \small
    \begingroup
    \setlength{\tabcolsep}{8pt}
    \renewcommand{\arraystretch}{1.15}
    \begin{tabular}{@{}llrr@{}}
      \toprule
      Attribute & Method & Acc $\uparrow$ & Balanced Acc $\uparrow$ \\
      \midrule
      Turbidity & Linear probe & \textbf{83.5} & \textbf{82.0} \\
      Turbidity & KNN probe & 79.0 & 73.5 \\
      Color cast & Linear probe & \textbf{89.0} & \textbf{89.7} \\
      Color cast & KNN probe & 78.5 & 79.6 \\
      \bottomrule
    \end{tabular}
    \endgroup
  \end{table}

  \subsection[Expert Review and Inter-Annotator Agreement]{\texorpdfstring{{Expert Review and Inter-Annotator Agreement}}{Expert Review and Inter-Annotator Agreement}}
  \label{app:expert-review-iaa}
  \par
  We report inter-annotator agreement and the LMM annotation error rate over 934 attribute annotations reviewed by three marine-biology master's students. The majority agreement rate is 93.15\%, the strict three-way agreement rate is 63.92\%, the complete-disagreement rate is 6.85\%, and the average pairwise agreement is 73.66\%. Expert review identifies 43 attributes requiring correction, corresponding to an annotation error rate of 4.60\%. These metrics are calculated as follows:
  \[
  \begin{aligned}
  A_{\mathrm{majority}} &= \frac{n_{\geq 2}}{N}\times100\%,\\
  A_{\mathrm{three\text{-}way}} &= \frac{n_3}{N}\times100\%,\\
  D_{\mathrm{complete}} &= \frac{n_0}{N}\times100\%,\\
  A_{\mathrm{pairwise}} &= \frac{n_{AB}+n_{AC}+n_{BC}}{3N}\times100\%,\\
  E_{\mathrm{annotation}} &= \frac{n_{\mathrm{corrected}}}{N}\times100\%.
  \end{aligned}
  \]
  Here, \(N=934\) is the number of reviewed attribute annotations; \(n_{\geq 2}=870\), \(n_3=597\), and \(n_0=64\) count annotations with agreement among at least two annotators, all three annotators, and no annotator pair, respectively. \(n_{AB}=810\), \(n_{AC}=657\), and \(n_{BC}=597\) are the agreement counts for the three annotator pairs, and \(n_{\mathrm{corrected}}=43\) is the number of attributes corrected during expert review.
  \par

  \section{Counting and Density-Evidence Benchmark}
  \label{app:counting-benchmark}

  This section provides additional details for Table~\ref{tab:unified-counting-results}. The benchmark is organized into two evaluation tracks. Track A uses only RGB images and evaluates whether a model can estimate fish abundance, localize the densest area, and reason about camouflage or visibility. Track B uses RGB images together with density heatmap evidence and evaluates whether a model can read density evidence, localize hotspots, and produce grounded ecological interpretations. The main paper reports a compact version of these results; here we provide the expanded transfer comparison, result interpretation, and metric definitions.

  \subsection{Expanded Nautilus Transfer Comparison}
  \label{app:expanded-nautilus-transfer}

  Table~\ref{tab:appendix-nautilus-transfer} expands the Nautilus transfer result summarized in Panel C of Table~\ref{tab:unified-counting-results}. All models are Qwen2.5-VL-7B variants evaluated on the same Track A test split with 1{,}129 counting samples. The Nautilus row uses Nautilus counting supervision, while the Nereus rows use different subsets of our grounded counting supervision.

  \begin{table}[H]
    \caption{Expanded transfer comparison on the Track A counting test split. Nereus variants are compared with Nautilus counting SFT under the same Qwen2.5-VL-7B backbone.}
    \label{tab:appendix-nautilus-transfer}
    \centering
    \scriptsize
    \setlength{\tabcolsep}{3.0pt}
    \renewcommand{\arraystretch}{0.95}
    \resizebox{\linewidth}{!}{%
    \begin{tabular}{@{}lrrrrrrrr@{}}
      \toprule
      Training data / setting & MAE $\downarrow$ & RMSE $\downarrow$ & Exact $\uparrow$ & \crmetric{5} $\uparrow$ & \crmetric{10} $\uparrow$ & Densest Cell $\uparrow$ & Camou. F1 $\uparrow$ & Count \\
      \midrule
      Nautilus counting SFT & 39.96 & 158.50 & 9.65 & 16.30 & 24.53 & 6.11 & 8.48 & 1129 \\
      Nereus count-only SFT & 38.57 & 89.59 & 9.92 & 16.83 & 26.66 & 6.11 & 7.41 & 1129 \\
      Nereus QA+structured-final SFT & \textbf{31.98} & \textbf{78.25} & 10.01 & \textbf{18.51} & \textbf{29.85} & \textbf{49.16} & 52.85 & 1129 \\
      Nereus full SFT & 36.19 & 84.74 & \textbf{10.36} & 16.56 & 25.33 & 47.39 & \textbf{54.06} & 1129 \\
      \bottomrule
    \end{tabular}}
  \end{table}

  \begin{table}[H]
    \caption{Relative changes over Nautilus counting SFT. Negative values are better for MAE and RMSE; positive values are better for all accuracy and F1 metrics.}
    \label{tab:appendix-nautilus-delta}
    \centering
    \scriptsize
    \setlength{\tabcolsep}{4.0pt}
    \renewcommand{\arraystretch}{0.95}
    \resizebox{\linewidth}{!}{%
    \begin{tabular}{@{}lrrrrrr@{}}
      \toprule
      Nereus setting & $\Delta$ MAE & $\Delta$ RMSE & $\Delta$ \crmetric{5} & $\Delta$ \crmetric{10} & $\Delta$ Densest Cell & $\Delta$ Camou. F1 \\
      \midrule
      Count-only SFT & -1.39 & -68.91 & +0.53 & +2.13 & +0.00 & -1.08 \\
      QA+structured-final SFT & \textbf{-7.98} & \textbf{-80.25} & \textbf{+2.21} & \textbf{+5.31} & \textbf{+43.05} & +44.37 \\
      Full SFT & -3.77 & -73.76 & +0.27 & +0.80 & +41.28 & \textbf{+45.57} \\
      \bottomrule
    \end{tabular}}
  \end{table}

  Table~\ref{tab:internvl-image-only-ablation} reports the image-only ablation on InternVL2.5-8B with RGB input, isolating the effect of different counting supervision forms.

  \begin{table}[H]
    \caption{Image-only ablation on InternVL2.5-8B with RGB input. Best values are shown in bold.}
    \label{tab:internvl-image-only-ablation}
    \centering
    \scriptsize
    \setlength{\tabcolsep}{5.0pt}
    \renewcommand{\arraystretch}{0.95}
    \resizebox{0.88\linewidth}{!}{%
    \begin{tabular}{@{}lrrrrr@{}}
      \toprule
      Training Setting & MAE $\downarrow$ & Exact Acc. $\uparrow$ & Acc@$\pm$5\% $\uparrow$ & Densest Cell Acc. $\uparrow$ & Camouflage F1 $\uparrow$ \\
      \midrule
      Zero-shot & 87.38 & 2.22 & 2.22 & 22.85 & -- \\
      Prompted zero-shot & 167.55 & 0.00 & 1.16 & 20.73 & 0.00 \\
      Count-only SFT & 31.72 & 10.01 & 19.04 & 20.81 & -- \\
      QA+structured-final SFT & \textbf{25.99} & \textbf{10.45} & \textbf{20.46} & \textbf{52.70} & 50.02 \\
      Full Count SFT & 30.48 & 9.74 & 18.95 & 51.90 & \textbf{52.59} \\
      \bottomrule
    \end{tabular}}
  \end{table}

  \subsection[Specialized Counting Baselines]{\texorpdfstring{{Specialized Counting Baselines}}{Specialized Counting Baselines}}
  \label{app:specialized-counting-baselines}
  \par
  We evaluate CountGD~\cite{countgd2024} and CLIP-Count~\cite{clipcount2023} as specialized counting baselines on the same Track A test split used in Table~\ref{tab:unified-counting-results}. Both models use their FSC-147-pretrained checkpoints in a zero-shot setting, with ``fish'' as the text query. We report numerical counting and densest-cell localization results; these models do not produce the camouflage-state, density-evidence, or ecology-grounded language outputs evaluated in the broader Nereus benchmark.
  \par

  \begin{table}[H]
    \caption[Specialized counting baselines on the Track A test split.]{{Specialized counting baselines on the Track A test split. Best values among the two specialized baselines are shown in bold.}}
    \label{tab:specialized-counting-baselines}
    \centering
    \scriptsize
    \begingroup
    \setlength{\tabcolsep}{4pt}
    \renewcommand{\arraystretch}{1.1}
    \begin{tabular}{@{}llrrrrrr@{}}
      \toprule
      Backbone & Setting & MAE $\downarrow$ & RMSE $\downarrow$ & Exact $\uparrow$ & \crmetric{5} $\uparrow$ & \crmetric{10} $\uparrow$ & Densest Cell $\uparrow$ \\
      \midrule
      CountGD & ZS & \textbf{32.33} & \textbf{80.16} & \textbf{6.64} & \textbf{14.35} & \textbf{25.69} & \textbf{67.67} \\
      CLIP-Count & ZS & 81.20 & 149.04 & 1.06 & 3.90 & 7.79 & 33.22 \\
      \bottomrule
    \end{tabular}
    \endgroup
  \end{table}

  \par
  CountGD achieves lower MAE and higher densest-cell accuracy than the zero-shot VLMs in Table~\ref{tab:unified-counting-results}. With Nereus supervision, Qwen2.5-VL-7B and InternVL2.5-8B achieve comparable or lower MAE, while CountGD retains the highest densest-cell accuracy among the evaluated models. These results highlight the complementary strengths of specialized counting models and Nereus-trained VLMs, which jointly support counting, spatial localization, camouflage reasoning, density interpretation, and ecological grounding.
  \par

  \subsection{Result Interpretation}
  \label{app:result-interpretation}

  The expanded results show that Nautilus counting supervision transfers to the Nereus test split, but mainly improves numerical counting and provides limited support for spatial localization or camouflage reasoning. Nereus count-only SFT obtains a similar MAE to Nautilus counting SFT, but its RMSE is much lower, indicating fewer severe outlier predictions.

  The QA+structured-final setting is the strongest overall Track A setting for counting and spatial evidence: it achieves the lowest MAE and RMSE, the best {Acc@{$\pm$}5\% and Acc@{$\pm$}10\%}, and a large improvement in Densest Cell accuracy. This suggests that converting point annotations into structured region and grid supervision helps the model learn where organisms are concentrated, not only how many are present. The full Nereus setting obtains the best Camouflage F1, showing that complete supervision is especially useful for visibility and camouflage-aware reasoning. The small trade-off in MAE relative to QA+structured-final SFT suggests that different supervision mixtures emphasize different parts of the task.

  For Track B in Table~\ref{tab:unified-counting-results}, density QA supervision substantially improves density-evidence reasoning for open-source models. For example, Qwen2.5-VL-7B improves from 18.13 to 45.26 in Density F1 and from 16.04 to 37.19 in Grounding F1, while InternVL2.5-8B shows a similar gain. LLaVA-OneVision-7B reaches the best Hotspot accuracy under full SFT, but its density and grounding F1 remain lower than Qwen and InternVL. These results indicate that reading a heatmap, localizing the hotspot, and explaining the density evidence are related but not identical capabilities.

  \subsection{Qualitative QA Examples}
  \label{app:qualitative-qa-examples}

\begin{figure}[H]
    \centering
    \includegraphics[width=\linewidth]{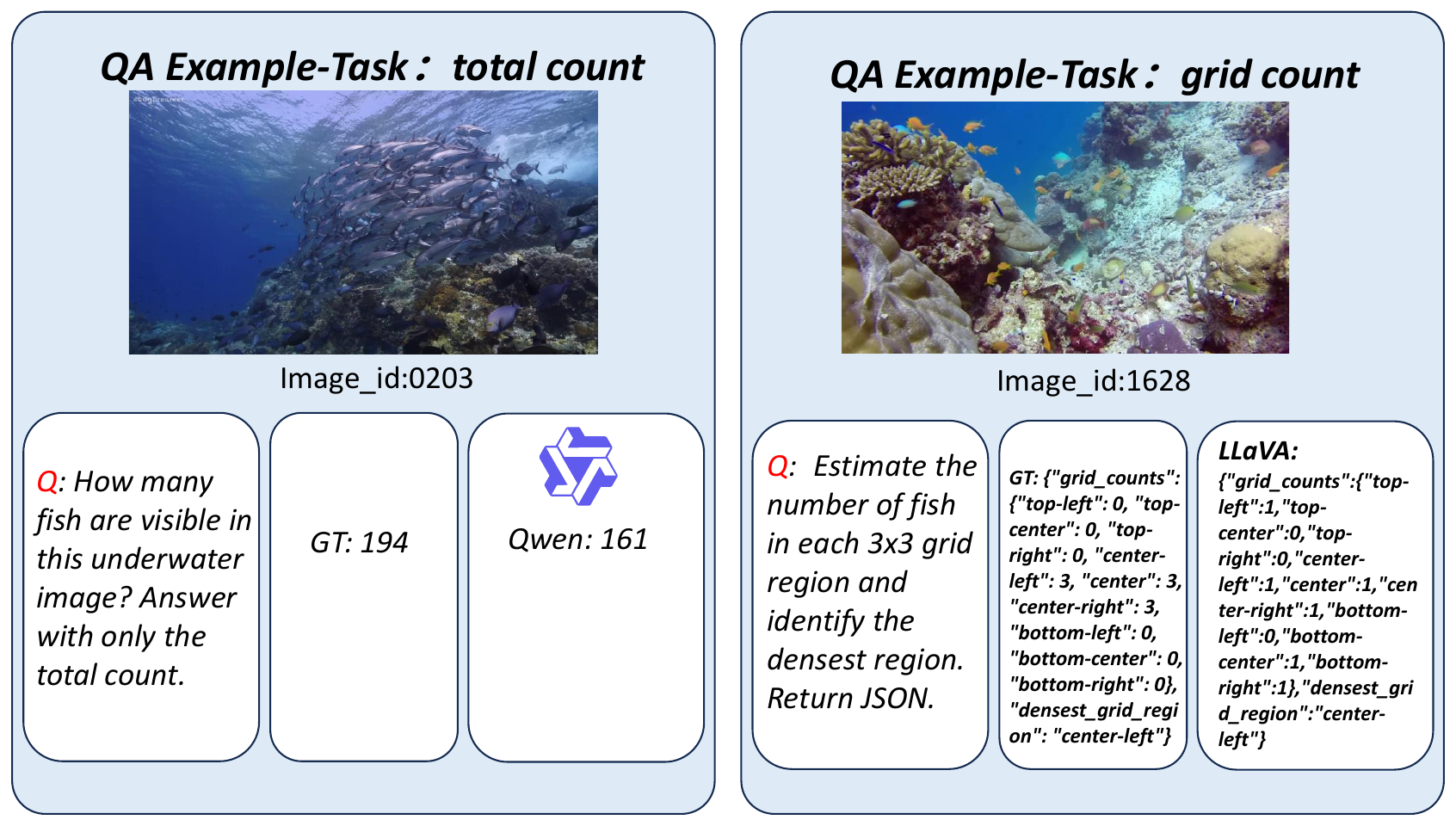}
    \caption{Qualitative examples for total-count and grid-count QA. The examples compare model outputs with grounded counting annotations, including total fish count and densest-region prediction.}
    \label{fig:appendix-qa-count-grid}
  \end{figure}

\begin{figure}[H]
    \centering
    \includegraphics[width=\linewidth]{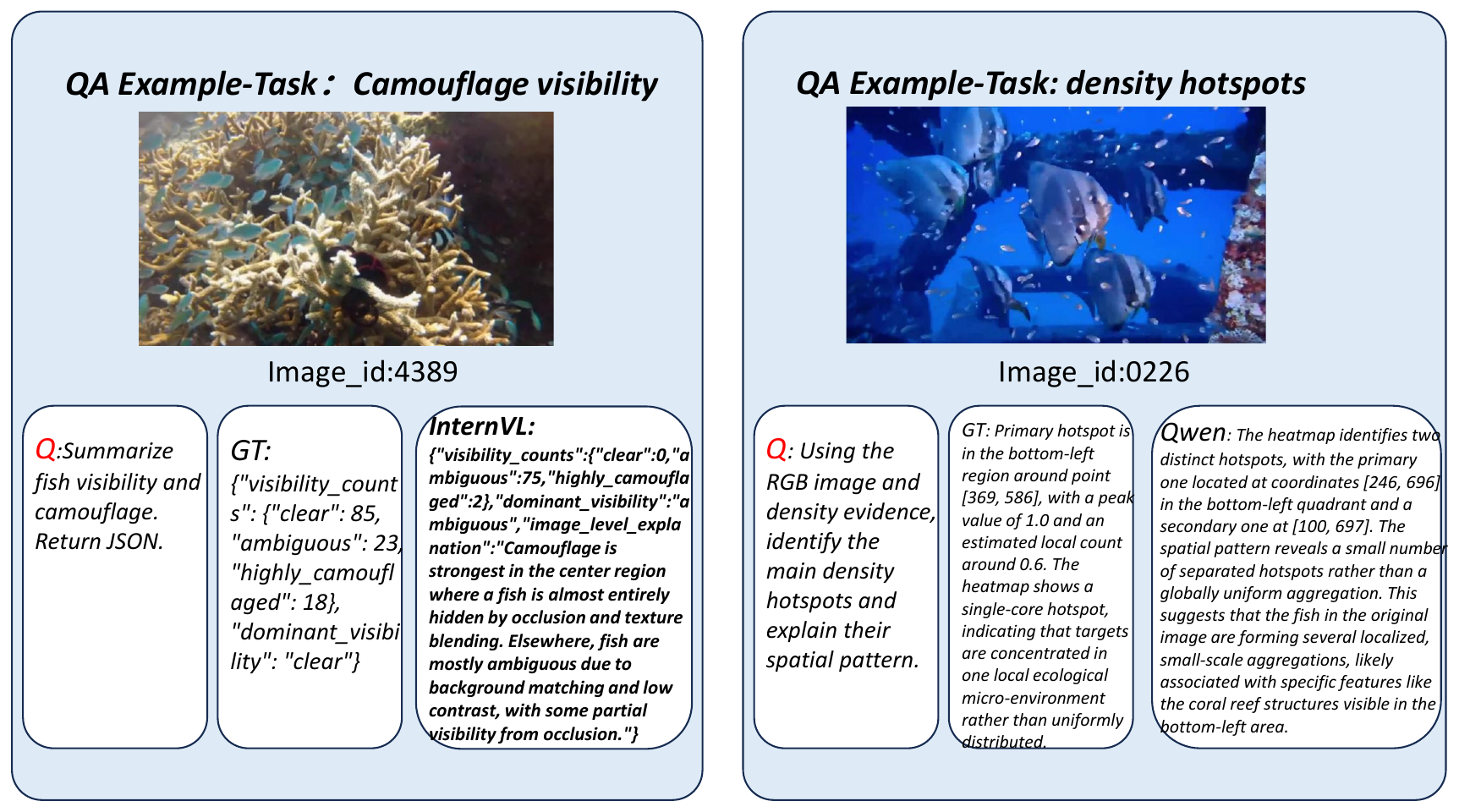}
    \caption{Qualitative examples for camouflage visibility and density-hotspot QA. The examples show visibility-state counting and hotspot reasoning from RGB images and density evidence.}
    \label{fig:appendix-qa-camouflage-hotspot}
  \end{figure}

\begin{figure}[H]
    \centering
    \includegraphics[width=\linewidth]{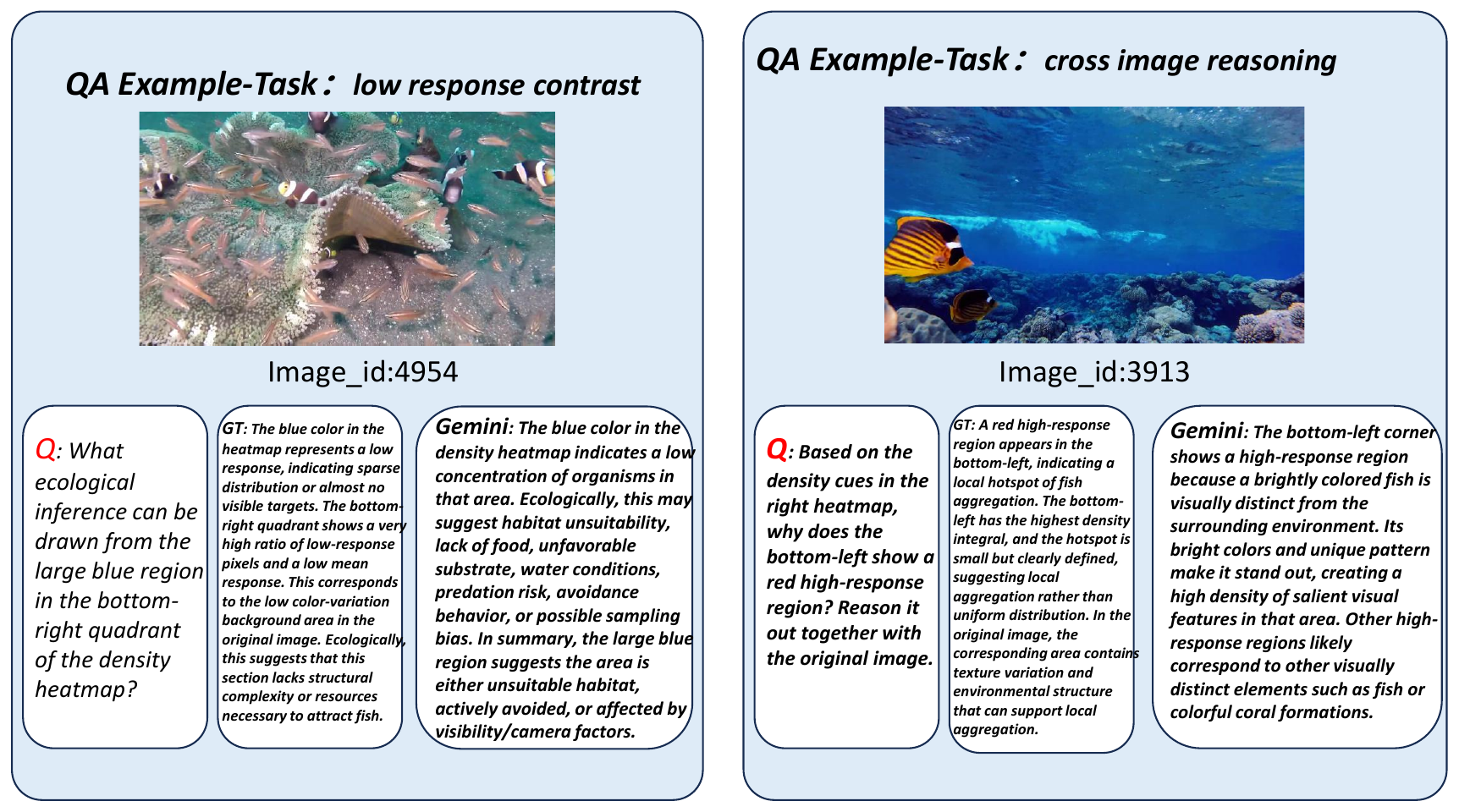}
    \caption{Qualitative examples for density-response interpretation. The examples cover ecological inference from low-response heatmap regions and cross-image reasoning over high-response density cues.}
    \label{fig:appendix-qa-density-reasoning}
  \end{figure}

  Figures~\ref{fig:appendix-qa-count-grid}--\ref{fig:appendix-qa-density-reasoning} show representative QA examples from the counting and density-evidence benchmark. These cases illustrate how the benchmark evaluates total counting, grid-based spatial counting, camouflage visibility, density hotspot localization, low-response interpretation, and cross-image density reasoning.

  \subsection{Metric Definitions}
  \label{app:metric-definitions}

  Let $c_i$ and $\hat{c}_i$ denote the ground-truth and predicted counts for image $i$, and let $N$ be the number of evaluated samples. All accuracy and F1 values are reported as percentages in the tables. Lower values are better for MAE and RMSE, while higher values are better for all other metrics.

  {\scriptsize
  \setlength{\tabcolsep}{2.5pt}
  \renewcommand{\arraystretch}{0.88}
  \begin{longtable}{@{}p{0.17\linewidth}p{0.08\linewidth}p{0.35\linewidth}p{0.31\linewidth}@{}}
    \caption{Metrics used in the counting and density-evidence benchmark.}
    \label{tab:appendix-counting-metrics}\\
    \toprule
    Metric & Track & Computation & Interpretation \\
    \midrule
    \endfirsthead
    \toprule
    Metric & Track & Computation & Interpretation \\
    \midrule
    \endhead
    Count MAE $\downarrow$ & A &
    $\frac{1}{N}\sum_i |\hat{c}_i-c_i|$ &
    Average absolute counting error; the primary RGB counting metric. \\
    Count RMSE $\downarrow$ & A &
    $\sqrt{\frac{1}{N}\sum_i(\hat{c}_i-c_i)^2}$ &
    Count error with stronger penalty for severe over-counting or under-counting. \\
    Exact Acc $\uparrow$ & A &
    $\frac{1}{N}\sum_i \mathbb{1}[\hat{c}_i=c_i]$ &
    Strict exact-count accuracy; any count deviation is incorrect. \\
    \crmetric{5} $\uparrow$ & A &
    $\frac{1}{N}\sum_i \mathbb{1}[|\hat{c}_i-c_i|\leq 0.05c_i]$ &
    {Count accuracy within {$\pm$}5\% of the ground-truth count.} \\
    \crmetric{10} $\uparrow$ & A &
    $\frac{1}{N}\sum_i \mathbb{1}[|\hat{c}_i-c_i|\leq 0.10c_i]$ &
    {Count accuracy within {$\pm$}10\% of the ground-truth count.} \\
    Densest Cell Acc $\uparrow$ & A &
    $\frac{1}{N}\sum_i \mathbb{1}[\hat{r}_i=r_i^\star]$ &
    Accuracy of predicting the densest cell in a $3\times3$ grid; evaluates spatial counting. \\
    Camouflage F1 $\uparrow$ & A &
    $2PR/(P+R)$ over camouflage-related answers &
    F1 for visibility and camouflage reasoning over clear, ambiguous, and highly camouflaged states. \\
    Hotspot Acc $\uparrow$ & B &
    $\frac{1}{N}\sum_i \mathbb{1}[\hat{h}_i=h_i^\star]$ &
    Accuracy of locating the density hotspot from RGB plus heatmap evidence. \\
    Density F1 $\uparrow$ & B &
    F1 over normalized density-reasoning concepts &
    F1 over density concepts such as hotspot, low-density area, aggregation, and dispersion. \\
    Ecology/Grounding F1 $\uparrow$ & B &
    F1 over ecological and grounding evidence units &
    F1 for ecological explanations grounded in image and density evidence. \\
    \bottomrule
  \end{longtable}
  }

  {For a ground-truth count of 3, the {$\pm$}5\% tolerance is 0.15, so predictions of 0 or 6 are both incorrect. For integer predictions and ground-truth counts below 20, Acc@{$\pm$}5\% is equivalent to exact matching.}

  \par
  The 1,129 Track-A evaluation samples span a range of ground-truth counts, as summarized in Table~\ref{tab:track-a-count-distribution}.
  \par

  \begin{table}[H]
    \caption[Ground-truth count distribution of the 1,129 Track-A evaluation samples.]{{Ground-truth count distribution of the 1,129 Track-A evaluation samples.}}
    \label{tab:track-a-count-distribution}
    \centering
    \small
    \begingroup
    \setlength{\tabcolsep}{12pt}
    \renewcommand{\arraystretch}{1.15}
    \begin{tabular}{@{}lrr@{}}
      \toprule
      Ground-truth count & Samples & Proportion \\
      \midrule
      0--50 & 526 & 46.59\% \\
      51--100 & 204 & 18.07\% \\
      101--200 & 183 & 16.21\% \\
      $>200$ & 216 & 19.13\% \\
      \bottomrule
    \end{tabular}
    \endgroup
  \end{table}

  Overall, Track A should be read as a test of RGB-only counting, spatial localization, and camouflage understanding. Track B should be read as a test of whether a model can use explicit density evidence to localize hotspots and produce reliable grounded explanations.

  \Needspace{28\baselineskip}
  \subsection[Task Composition of QAS and Full]{\texorpdfstring{{Task Composition of QAS and Full}}{Task Composition of QAS and Full}}
  \label{app:qas-full-composition}
  \par
  QAS contains 47,340 Track-A training records derived from 3,945 images. Full retains all QAS records and adds 23,655 Track-B records, yielding 70,995 records in total. Table~\ref{tab:qas-full-composition} reports the task composition and natural record-level mixture before shuffling; Track A and Track B account for 66.68\% and 33.32\% of Full, respectively.
  \par

  \begin{table}[H]
    \caption[Task composition and record-level proportions of QAS and Full.]{{Task composition and record-level proportions of QAS and Full.}}
    \label{tab:qas-full-composition}
    \centering
    \small
    \begingroup
    \setlength{\tabcolsep}{5pt}
    \renewcommand{\arraystretch}{1.15}
    \begin{tabular}{@{}llcrr@{}}
      \toprule
      Setting & Component & Track & Records & \shortstack{Ratio within\\setting} \\
      \midrule
      QAS & Region/grid counting QA & A & 15,780 & 33.33\% \\
      QAS & Camouflage QA & A & 15,780 & 33.33\% \\
      QAS & Structured total count & A & 3,945 & 8.33\% \\
      QAS & Structured grid count & A & 3,945 & 8.33\% \\
      QAS & Structured camouflage & A & 3,945 & 8.33\% \\
      QAS & Structured final summary & A & 3,945 & 8.33\% \\
      \textbf{QAS total} & & & \textbf{47,340} & \textbf{100\%} \\
      \midrule
      Full & All QAS components & A & 47,340 & 66.68\% \\
      Full & Density QA & B & 15,765 & 22.21\% \\
      Full & Structured density hotspot & B & 3,945 & 5.56\% \\
      Full & Structured ecological grounding & B & 3,945 & 5.56\% \\
      \textbf{Full total} & & & \textbf{70,995} & \textbf{100\%} \\
      \bottomrule
    \end{tabular}
    \endgroup
  \end{table}

  \clearpage
  \section[Data Splits and Additional Evaluation]{\texorpdfstring{{Data Splits and Additional Evaluation}}{Data Splits and Additional Evaluation}}
  \label{app:split-additional-evaluation}

  \subsection[Official IOCfish5K Protocol]{\texorpdfstring{{Official IOCfish5K Protocol}}{Official IOCfish5K Protocol}}
  \label{app:official-iocfish-protocol}
  \par
  Table~\ref{tab:unified-counting-results} uses a Nereus-specific split for unified counting and density-grounded evaluation. All models in Table~\ref{tab:unified-counting-results} share this split, which differs from the official IOCfish5K protocol. We provide the official split definitions and an image-ID mapping utility to associate Nereus annotations and QA with the official splits.

  Under the official protocol, we evaluate the public IOCFormer checkpoint and InternVL2.5-8B QAS, trained using the official 3,137/500-image train/validation split, on the same 2,000-image test set with a shared evaluator.
  \par

  \begin{table}[H]
    \caption[Comparison under the official IOCfish5K protocol.]{{Comparison under the official IOCfish5K protocol.}}
    \label{tab:official-iocfish-protocol}
    \centering
    \scriptsize
    \begingroup
    \setlength{\tabcolsep}{3pt}
    \renewcommand{\arraystretch}{1.2}
    \resizebox{\linewidth}{!}{%
    \begin{tabular}{@{}llrrrrrr@{}}
      \toprule
      Model & Setting & MAE $\downarrow$ & RMSE $\downarrow$ & Exact $\uparrow$ & Acc@{$\pm$}5\% $\uparrow$ & Acc@{$\pm$}10\% $\uparrow$ & Failure $\downarrow$ \\
      \midrule
      IOCFormer & Public checkpoint & \textbf{17.52} & \textbf{43.13} & 7.45 & \textbf{22.95} & \textbf{39.95} & 0 \\
      InternVL2.5-8B QAS & Official train/validation split & 31.16 & 82.31 & \textbf{10.65} & 19.50 & 30.45 & 0 \\
      \bottomrule
    \end{tabular}}
    \endgroup
  \end{table}

  \clearpage
  \section{NeurIPS Paper Checklist}
\label{app:checklist}

\begin{enumerate}

\item {\bf Claims}
    \item[] Question: Do the main claims made in the abstract and introduction accurately reflect the paper's contributions and scope?
    \item[] Answer: \answerYes{} 
    \item[] Justification: The abstract and Introduction state the dataset scope, task formulation, annotation types, and benchmark claims, which are supported by the dataset statistics and experimental results in Sections 3--5.


\item {\bf Limitations}
    \item[] Question: Does the paper discuss the limitations of the work performed by the authors?
    \item[] Answer: \answerYes{} 
    \item[] Justification: {We discuss limitations and future directions in Section~{\ref{sec:limitations}}.}

\item {\bf Theory assumptions and proofs}
    \item[] Question: For each theoretical result, does the paper provide the full set of assumptions and a complete (and correct) proof?
    \item[] Answer: \answerNA{} 
    \item[] Justification: The paper does not present theoretical results, theorems, or formal proofs.


    \item {\bf Experimental result reproducibility}
    \item[] Question: Does the paper fully disclose all the information needed to reproduce the main experimental results of the paper to the extent that it affects the main claims and/or conclusions of the paper (regardless of whether the code and data are provided or not)?
    \item[] Answer: \answerYes{}
    \item[] Justification: We explain our implementation details in Section~\ref{sec:experiment}, and provide more details in Appendix~\ref{app:training-details}.


\item {\bf Open access to data and code}
    \item[] Question: Does the paper provide open access to the data and code, with sufficient instructions to faithfully reproduce the main experimental results, as described in supplemental material?
    \item[] Answer: \answerYes{}
    \item[] Justification: {We provide the dataset annotations, code, licenses, and instructions for data access and reproducing the main experiments through the link on the first page of the paper.}


\item {\bf Experimental setting/details}
    \item[] Question: Does the paper specify all the training and test details (e.g., data splits, hyperparameters, how they were chosen, type of optimizer) necessary to understand the results?
    \item[] Answer: \answerYes{} 
    \item[] Justification: We explain these details in the Implementation Details section and Appendix~\ref{app:training-details}.

\item {\bf Experiment statistical significance}
    \item[] Question: Does the paper report error bars suitably and correctly defined or other appropriate information about the statistical significance of the experiments?
    \item[] Answer: \answerNo{}
    \item[] Justification: The paper reports benchmark results without error bars or statistical significance tests, mainly due to the computational cost of fine-tuning multiple large multimodal models across repeated runs.

\item {\bf Experiments compute resources}
    \item[] Question: For each experiment, does the paper provide sufficient information on the computer resources (type of compute workers, memory, time of execution) needed to reproduce the experiments?
    \item[] Answer: \answerYes{} 
    \item[] Justification: Section 5 and Appendix~\ref{app:training-details} report the main compute setup. The automated annotation pipeline uses four RTX 4090 GPUs, and the comparison and ablation fine-tuning experiments are conducted on four RTX 4090 GPUs over 12 days in total.
    
\item {\bf Code of ethics}
    \item[] Question: Does the research conducted in the paper conform, in every respect, with the NeurIPS Code of Ethics \url{https://neurips.cc/public/EthicsGuidelines}?
    \item[] Answer: \answerYes{}
    \item[] Justification: The work uses underwater ecological imagery and annotation resources with license-aware processing, and does not involve sensitive personal data or human-subject experiments.


\item {\bf Broader impacts}
    \item[] Question: Does the paper discuss both potential positive societal impacts and negative societal impacts of the work performed?
    \item[] Answer: \answerYes{} 
    \item[] Justification: Appendix~\ref{app:broader-impacts} discusses positive uses for marine research, conservation, monitoring, and education, as well as risks from domain shift, erroneous ecological interpretation, misuse for resource extraction, and sensitive location disclosure.
    
\item {\bf Safeguards}
    \item[] Question: Does the paper describe safeguards that have been put in place for responsible release of data or models that have a high risk for misuse (e.g., pre-trained language models, image generators, or scraped datasets)?
    \item[] Answer: \answerNA{}
    \item[] Justification: The paper introduces an underwater perception dataset and does not release high-risk generative models or sensitive personal data.


\item {\bf Licenses for existing assets}
    \item[] Question: Are the creators or original owners of assets (e.g., code, data, models), used in the paper, properly credited and are the license and terms of use explicitly mentioned and properly respected?
    \item[] Answer: \answerYes{} 
    \item[] Justification: Appendix~\ref{app:license-details} discusses the licenses of existing data sources and explains how original image redistribution is handled.

\item {\bf New assets}
    \item[] Question: Are new assets introduced in the paper well documented and is the documentation provided alongside the assets?
    \item[] Answer: \answerYes{}
    \item[] Justification: The paper documents the new Nereus dataset, including data scale, annotation types, task families, degradation labels, and intended release protocol.


\item {\bf Crowdsourcing and research with human subjects}
    \item[] Question: For crowdsourcing experiments and research with human subjects, does the paper include the full text of instructions given to participants and screenshots, if applicable, as well as details about compensation (if any)? 
    \item[] Answer: \answerNA{} 
    \item[] Justification: The paper does not involve crowdsourcing nor research with human subjects.

\item {\bf Institutional review board (IRB) approvals or equivalent for research with human subjects}
    \item[] Question: Does the paper describe potential risks incurred by study participants, whether such risks were disclosed to the subjects, and whether Institutional Review Board (IRB) approvals (or an equivalent approval/review based on the requirements of your country or institution) were obtained?
    \item[] Answer: \answerNA{} 
    \item[] Justification: The paper does not involve research with human subjects.

\item {\bf Declaration of LLM usage}
    \item[] Question: Does the paper describe the usage of LLMs if it is an important, original, or non-standard component of the core methods in this research? Note that if the LLM is used only for writing, editing, or formatting purposes and does \emph{not} impact the core methodology, scientific rigor, or originality of the research, declaration is not required.
    \item[] Answer: \answerYes{}
    \item[] Justification: Section 4 describes the use of LMMs/VLMs in the annotation pipeline, including dual-model discussion, arbitration, knowledge-prior prompting, and grounded counting QA generation. Appendix~\ref{app:kg-prior-example} provides an example of the knowledge-prior prompt used during annotation.


\end{enumerate}

  \end{document}